\documentclass[11pt]{article}

\usepackage[final]{acl}

\usepackage{times}
\usepackage{latexsym}

\usepackage[T1]{fontenc}

\usepackage{microtype}

\usepackage{inconsolata}

\usepackage{graphicx}

\title{UrduFactCheck: An Agentic Fact-Checking Framework for Urdu \\with Evidence Boosting and Benchmarking}

\author{\textbf{Sarfraz Ahmad}\thanks{\xspace\xspace Equal contribution.} \quad  
        \textbf{Hasan Iqbal}$^*$ \quad 
        \textbf{Momina Ahsan} \quad
        \textbf{Numaan Naeem} \\  
        \textbf{Muhammad Ahsan Riaz Khan} \quad
        \textbf{Arham Riaz} \quad
        \textbf{Muhammad Arslan Manzoor} \\
        \textbf{Yuxia Wang} \quad
        \textbf{Preslav Nakov} \\
MBZUAI \quad \\
\texttt{\{sarfraz.ahmad, hasan.iqbal, preslav.nakov\}@mbzuai.ac.ae}
}

\usepackage{xspace}

\newcommand{\model}[1]{\textsc{#1}\xspace}

\newcommand{\gptfouro}{\model{GPT-4o}}
\newcommand{\gptfouromini}{\model{GPT-4o-mini}}

\newcommand{\gptfourpointone}{\model{GPT-4.1}}
\newcommand{\gptfourpointonemini}{\model{GPT-4.1-mini}}

\newcommand{\geminiOneFivePro}{\model{Gemini-1.5-pro}}
\newcommand{\claudesonnet}{\model{Claude-Sonnet}}
\newcommand{\claudeSonnetThreeFive}{\model{Claude-Sonnet-3.5}}
\newcommand{\claudehaiku}{\model{Claude-Haiku}}

\newcommand{\framework}[1]{\textsc{#1}\xspace}
\newcommand{\urdufactcheck}[1][]{\framework{UrduFactCheck#1}}

\newcommand{\factscore}{\framework{FactScore}}
\newcommand{\factcheckgpt}{\framework{FactCheckGPT}}
\newcommand{\factool}{\framework{Factool}}

\newcommand{\fire}{\framework{FIRE}}
\newcommand{\simpleqa}{\framework{SimpleQA}}
\newcommand{\openfactcheck}{\framework{OpenFactCheck}}

\newcommand{\loki}{\framework{Loki}}

\newcommand{\dataset}[1]{\textsc{#1}\xspace}

\newcommand{\freshqa}{\dataset{FreshQA}}

\newcommand{\factoolqa}{\dataset{FacTool-QA}}

\newcommand{\factcheckbench}{\dataset{Factcheck-Bench}}

\newcommand{\bingcheck}{\dataset{BingCheck}}
\newcommand{\urdufactqa}{\dataset{UrduFactQA}}
\newcommand{\urdufactbench}{\dataset{UrduFactBench}}

\newcommand{\module}[1]{\textsc{#1}\xspace}

\newcommand{\claimprocessor}{\module{ClaimProcessor}}
\newcommand{\retriever}{\module{Retriever}}
\newcommand{\verifier}{\module{Verifier}}
\newcommand{\querygenerator}{\module{QueryGenerator}}

\usepackage[utf8]{inputenc}     
\usepackage[T1]{fontenc}        
\usepackage{multirow}           
\usepackage{multicol}           
\usepackage{array}              
\usepackage{rotating}           
\usepackage{tikz}               
\usepackage{collcell}           
\usepackage{etoolbox}           
\usepackage{pgf}                
\usepackage{enumitem}           
\usepackage{adjustbox}
\usepackage{booktabs}
\usepackage{enumitem}
\usepackage{amsmath}
\usepackage{amssymb}

\usepackage{fontspec}
\usepackage[bidi=default]{babel}
\babelprovide[main, import]{english}
\babelprovide[import]{urdu}

\babelfont[urdu]{tt}[
  Path=fonts/,
  UprightFont = *,
  Script=Arabic,
  Language=Urdu,
  Scale=0.85
]{NotoNaskhArabic-Regular.ttf}
\babelfont[urdu]{rm}[
  Path=fonts/,
  UprightFont = *,
  Script=Arabic,
  Language=Urdu,
  Scale=0.85
]{NotoNastaliqUrdu-Regular.ttf}

\usepackage{caption}
\usepackage[most]{tcolorbox}
\tcbuselibrary{breakable}
\newtcolorbox{codebox}{
  breakable,
  enhanced,
  colback=black!2,
  colframe=black!20,
  left=0.5ex, right=0.5ex, top=1ex, bottom=1ex,
  sharp corners,
}

\newcommand{\tabref}[2][]{Table#1~\ref{#2}\xspace}
\newcommand{\figref}[1]{Figure~\ref{#1}\xspace}

\newcommand{\appref}[1]{Appendix~\ref{#1}\xspace}

\usepackage[table,x11names]{xcolor}
\usepackage{pgf}
\usepackage{xparse}
\usepackage{array,makecell,tabularx}

\newcommand*{\MinNumber}{0.0}%
\newcommand*{\MidNumber}{0.5}%
\newcommand*{\MaxNumber}{1.0}%

\ExplSyntaxOn
\NewDocumentCommand{\ApplyZeroOneGradient}{m}{\mygradient_apply:n{#1}}

\cs_new_protected:Npn \mygradient_apply:n #1
  {
    \regex_match:nnTF { \A \s* -? (?: \d+ (?:\.\d*)? | \.\d+ ) \s* \Z } {#1}
      {
        \pgfmathparse{#1}\let\val\pgfmathresult
        \pgfmathparse{\val > \MidNumber}\ifnum\pgfmathresult>0
          \pgfmathsetmacro{\PercentColor}{min(max(100*(\val-\MidNumber)/(\MaxNumber-\MidNumber),0),100)}
          \begingroup\setlength{\fboxsep}{0.3pt}%
            \colorbox{DarkSeaGreen1!\PercentColor!LightGoldenrod1}{\strut #1}%
          \endgroup
        \else
          \pgfmathsetmacro{\PercentColor}{min(max(100*(1 - (\val-\MinNumber)/(\MidNumber-\MinNumber)),0),100)}
          \begingroup\setlength{\fboxsep}{0.3pt}%
            \colorbox{LightPink1!\PercentColor!LightGoldenrod1}{\strut #1}%
          \endgroup
        \fi
      }
      {
        \begingroup\strut\detokenize{#1}\endgroup
      }
  }
\ExplSyntaxOff

\newcolumntype{L}[1]{>{\raggedright\let\newline\\\arraybackslash\hspace{0pt}}m{#1}}
\newcolumntype{C}[1]{>{\centering\let\newline\\\arraybackslash\hspace{0pt}}m{#1}}
\newcolumntype{R}[1]{>{\raggedleft\let\newline\\\arraybackslash\hspace{0pt}}m{#1}}
\newcolumntype{M}{>{\collectcell\ApplyZeroOneGradient}c<{\endcollectcell}}

\begin{document}
\maketitle
\begin{abstract}

The rapid adoption of Large Language Models (LLMs) has raised important concerns about the factual reliability of their outputs, particularly in low-resource languages such as Urdu. Existing automated fact-checking systems are predominantly developed for English, leaving a significant gap for the more than 200 million Urdu speakers worldwide. In this work, we present \urdufactbench and \urdufactqa, two novel hand-annotated benchmarks designed to enable fact-checking and factual consistency evaluation in Urdu. While \urdufactbench focuses on claim verification, \urdufactqa targets the factuality of LLMs in question answering. These resources, the first of their kind for Urdu, were developed through a multi-stage annotation process involving native Urdu speakers. To complement these benchmarks, we introduce \urdufactcheck, a modular fact-checking framework that incorporates both monolingual and translation-based evidence retrieval strategies to mitigate the scarcity of high-quality Urdu evidence. Leveraging these resources, we conduct an extensive evaluation of twelve LLMs and demonstrate that translation-augmented pipelines consistently enhance performance compared to monolingual ones. Our findings reveal persistent challenges for open-source LLMs in Urdu and underscore the importance of developing targeted resources. All code and data are publicly available at \url{https://github.com/mbzuai-nlp/UrduFactCheck}.

\end{abstract}

\section{Introduction}
\label{sec:introduction}

In recent years, the way we find and share information has changed dramatically. Large language models (LLMs) like \gptfouro~\cite{gpt-4o} are now capable of answering questions, generating articles, and even holding conversations that sound convincingly human. 

Despite all mentioned strengths, these models sometimes make mistakes and do so with surprising confidence, even when they’re wrong. This problem, known as ``hallucination''~\cite{bang2023multitask, borji2023categorical, tie2024llms}, is especially troubling when technology is used in important areas such as healthcare, finance, or law~\cite{chuang2023dola, geng-etal-2024-survey, wang-etal-2024-factuality}.

At the same time, social media platforms have become a main source of news and information for millions of people worldwide. These platforms are also a source of fake news and misinformation. During major events such as the 2016 U.S. Presidential Election and the Brexit referendum, false narratives were used to manipulate public opinion at scale~\cite{allcott2017social, pogue2017stamp, vosoughi2018spread}. The rapid algorithm driven spread of such content, especially on TikTok, Facebook, and Twitter, has reduced public trust in institutions and increased political polarization~\cite{zimmer2019fake, trilling2017newsworthiness}. This trend worsened during the COVID-19 pandemic, which not only increased public awareness of misinformation but also revealed its risks in real time. The World Health Organization (WHO) warned that we were facing not only a pandemic but also an ‘infodemic’, a surge of false or misleading information about the virus on social media~\cite{humprecht2020they, arechar2023understanding, world2023overview}.

\begin{figure*}[t!]
    \centering
    \includegraphics[width=\linewidth]{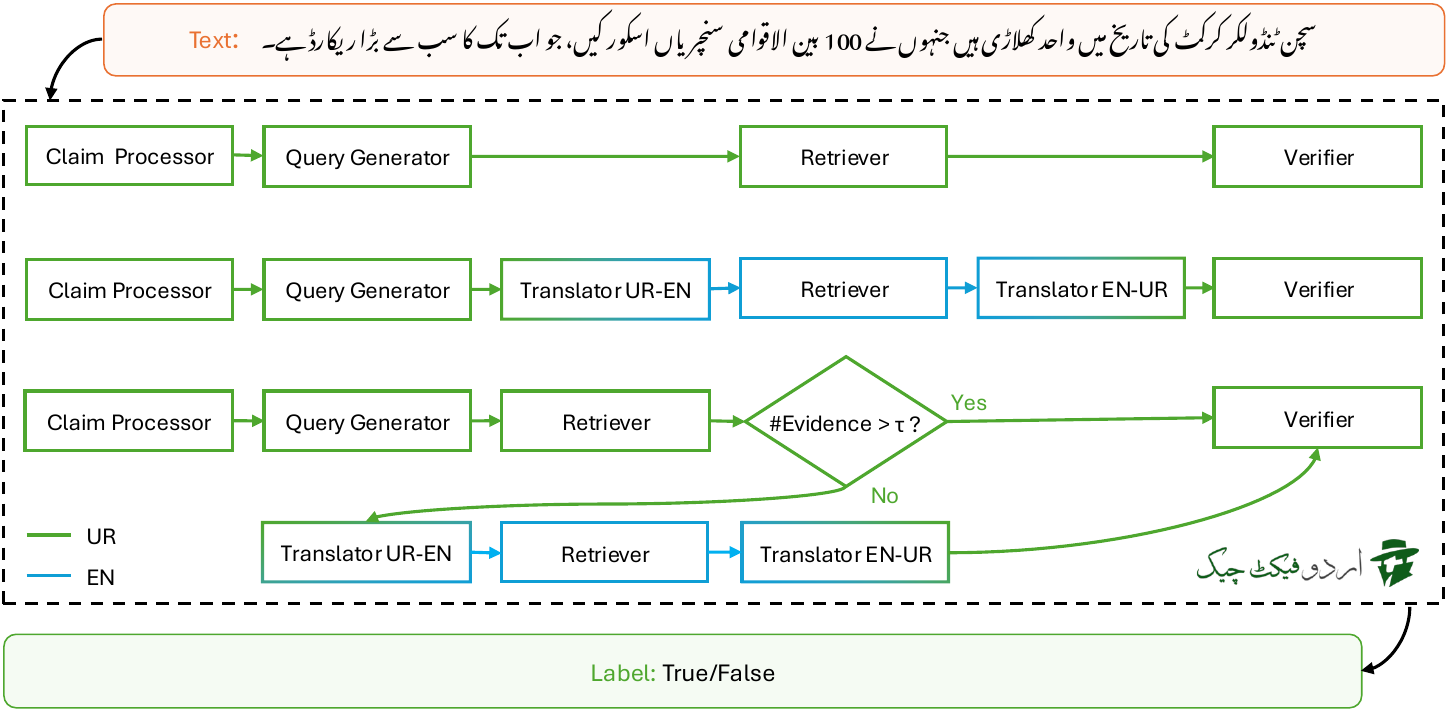}
    \caption{Three core fact-checking pipelines of \urdufactcheck: \textbf{(top)} end-to-end Urdu framework, \textbf{(middle)} translation-augmented retrieval, and \textbf{(bottom)} threshold-based rerouting using $\tau$ when No. of Evidence $< \tau$. Green indicates Urdu context, while blue indicates English context.}
    \label{fig:architecture}
\end{figure*}

Despite the growing momentum of fact-checking efforts, most initiatives focus on English-language content~\cite{guo2022survey}, leaving a gap for other widely spoken languages. Urdu is the national language of Pakistan, holds official status in several Indian states, and is spoken by about 232 million people worldwide, yet it accounts for less than 0.5\% of all online content.\footnote{\url{https://www.icls.edu/blog/most-spoken-languages-in-the-world}}

The development of automated fact-checking tools for Urdu remains limited, reflecting a broader trend in Natural Language Processing (NLP), where low-resource languages are often routinely left behind due to a lack of annotated data and low representation in LLM pretraining corpora~\cite{du2021cross, raja2023fake}. These gaps are concerning given the frequent spread of misinformation in Urdu on social media. False content circulates in various forms, including jokes, memes, and deliberate disinformation~\cite{amjad2022overview}. While there has been progress in applying cross-lingual methods to tasks such as hate speech and rumor detection in~\cite{glavavs2020xhate, haider2023detecting}, these advances have not extended to automated fact-checking in Urdu. To address this, we introduce three resources for factuality assessment in Urdu:

\paragraph{\textbf{\urdufactbench:}} 
A manually curated benchmark for claim verification, enabling the evaluation of fact-checking systems and comparisons to automated fact-checkers. It combines multiple claim datasets into a unified resource and standardizes their labels into a consistent format for evaluation.

\paragraph{\textbf{\urdufactqa:}}
A manually annotated dataset for evaluating the factual accuracy of LLMs on Urdu QA tasks. We use it to assess 12 state-of-the-art LLMs, and it serves as the first benchmark to measure the factuality of LLM answers in Urdu.

\paragraph{\textbf{\urdufactcheck:}} 
A fact-checking pipeline for claim verification and LLM evaluation in Urdu. It features a modular design with monolingual and translation-based retrieval, and uses a thresholded evidence boosting technique as illustrated in \figref{fig:architecture}.

\urdufactcheck builds on recent modular frameworks such as \loki~\cite{li-etal-2025-loki}, \openfactcheck~\cite{wang-etal-2025-openfactcheck, iqbal-etal-2024-openfactcheck}, and \fire~\cite{xie-etal-2025-fire}, and addresses key challenges in Urdu fact-checking:

\begin{itemize}
\item Detecting factual errors in free-form text;
\item Enhancing quality and coverage of evidence;
\item Evaluating factuality of LLMs on Urdu;
\item Analyzing system component contributions;
\end{itemize}

Our main contributions are two new Urdu datasets for factuality evaluation, supported by an open-source framework for systematic experimentation. These resources provide a foundation for future research on factuality in low-resource languages. In addition, they enable direct comparisons between proprietary and open-source LLMs on Urdu inputs. By making both datasets and framework publicly available, we aim to encourage reproducible research and facilitate extensions to other underrepresented languages.

\section{Related Work}
\label{sec:related-work}

Prior efforts in Urdu fact-checking have focused mainly on fake news classification. The \framework{UrduFake@FIRE2021}~\cite{amjad2022overview} shared task framed this as a binary problem, highlighting generalization issues under domain shifts. \framework{Ax-to-Grind}~\cite{harris2023ax} introduced a larger annotated dataset and applied multilingual models like \model{mBERT} and \model{XLNet}. More recently, \framework{Hook and Bait Urdu}~\cite{harris2025benchmarking} released the largest Urdu fake news corpus, leveraging LoRA fine-tuning of \model{LLaMA-2} for mono- and multilingual detection. While these systems improved classification and dataset scale, they do not provide end-to-end factuality pipelines.

In parallel, several Urdu QA datasets have emerged. \dataset{UQA}~\cite{arif2024uqa}, a span-preserving translation of \dataset{SQuAD2.0}~\cite{rajpurkar-etal-2018-know}, has been used to benchmark multilingual models like \model{mBERT} and \model{XLM-RoBERTa}. Other corpora such as \dataset{UQuAD1.0}~\cite{kazi2021uquad1} support extractive QA but do not assess factual correctness. These resources target reading comprehension and general QA, rather than fact-checking or systematic evaluation of generated responses.

Multilingual benchmarks such as \dataset{X-Fact}~\cite{gupta2021x} have tested LLM factuality across several low-resource languages, but Urdu remains underrepresented in these evaluations. At the same time, tools like \factscore~\cite{min2023factscore}, \factool~\cite{chern2023factool}, and \factcheckgpt~\cite{wang-etal-2024-factcheck} have advanced metrics, retrieval strategies, and modularity in fact-checking, but they are developed mainly for English and lack Urdu-specific datasets. This absence limits systematic evaluation of factuality in Urdu and restricts the transfer of existing methods to this language.

In contrast, this work introduces \urdufactbench\ and \urdufactqa, the first datasets for fact-checking and factuality evaluation in Urdu. \urdufactbench\ supports claim verification, while \urdufactqa\ evaluates the factual accuracy of LLM-generated responses. To support their use, we also provide \urdufactcheck, a modular framework with claim processing, evidence retrieval, and verification. Together, these resources provide the first foundation for factuality evaluation in Urdu, addressing gaps from prior work on classification or general QA.

\section{Datasets}
\label{sec:datasets}

To provide a foundation for evaluating automated fact-checkers and measuring the factuality of LLMs, we draw on three claim verification datasets and two QA datasets, curating them into \urdufactbench and \urdufactqa, respectively.

\subsection{Dataset Collection}

Given the limited availability of factual datasets in Urdu, we adapted established English datasets through a multi-stage process under expert supervision. For claim verification, we selected three datasets: \bingcheck~\cite{li2023self}, \factcheckbench~\cite{wang-etal-2024-factcheck}, and \factool~\cite{chern2023factool}. From \factool, which spans multiple domains, we extracted a subset requiring world knowledge, referred to as \factoolqa. 

For factual question answering, we used two QA datasets: \simpleqa~\cite{wei2024measuring} and \freshqa~\cite{vu-etal-2024-freshllms}. Together, these five datasets cover varied claim structures, domains, and formats, allowing evaluation of both claim-level verification and the factual behavior of LLMs. We emphasized claims that require external knowledge, as this is central to factuality tasks. For \factcheckbench\ and \bingcheck, we simplified the original four-label scheme (\textit{supported}, \textit{partially supported}, \textit{not supported}, \textit{refuted}) into binary: \textit{True} for \textit{supported}/\textit{partially supported}, \textit{False} for \textit{refuted}, and removal of \textit{not supported}. This standardization ensured consistency across datasets.

To reduce class imbalance in \bingcheck, which contained 3,581 \textit{True} and only 42 \textit{False} claims, we sampled 100 \textit{True} instances for the test set. This produced a more balanced dataset and enabled evaluation metrics to better capture performance across both classes.

\subsection{Translation and Annotation}

To begin the translation process, we used three LLMs: \gptfouro~\cite{gpt-4o}, \geminiOneFivePro~\cite{team2024gemini}, and \claudeSonnetThreeFive~\cite{claude35}. Each model was tested by translating 50 claims and 50 QA pairs in a few-shot setup, and the translations were reviewed by expert annotators for fluency, adequacy, and correctness. Based on these qualitative assessments by annotators, \gptfouro was selected as the most suitable model for translation. \\

This machine-generated translation was not intended as a final product, but rather as a way to accelerate the annotation workflow and reduce the manual workload for human annotators. By leveraging \gptfouro, annotators were able to dedicate their efforts to quality assurance, validation, and refinement rather than translating from scratch.

To improve translation quality, annotators created 100 demonstration examples (20 per dataset) for the LLM’s few-shot setup. A custom prompt template, guided by linguistic rules, instructed the model to generate grammatically correct Urdu while preserving technical terms and transliterations. The guidelines also addressed left-to-right numerals in right-to-left flow and correct placement of acronyms (see \appref{app:pre-translation}). For optimal few-shot performance, Max Marginal Relevance (MMR) was used to select the most relevant examples. The pipeline was implemented with \framework{LangChain}\footnote{\url{https://www.langchain.com}} and an output parser, using default OpenAI Library parameters for \gptfouro.

After translation, each dataset underwent dual annotation: one expert reviewed the output, and a second independently validated it for linguistic consistency, factual correctness, and cultural appropriateness. A custom annotation portal further streamlined review and verification (see \appref{app:annotator}). This workflow ensured all datasets met high standards of quality and reliability for factuality evaluation in Urdu. Native Urdu-speaking annotators were employed to ensure the highest linguistic and cultural quality in the datasets. All annotators were required to be senior high-school graduates at minimum, with higher educational qualifications preferred, and both parents being from and residing in Urdu-speaking regions. This careful selection process helped guarantee not only fluency but also deep cultural familiarity with the language. The final translated datasets resulted in the following two resources:

\paragraph{\textbf{\urdufactbench:}}
Comprising the claim datasets \bingcheck, \factoolqa, and \factcheckbench, this benchmark serves as the ground truth for evaluating the performance of automated fact-checkers in Urdu (see \tabref{tab-urdufactbench}).

\paragraph{\textbf{\urdufactqa:}}
Consisting of the QA datasets \simpleqa\ and \freshqa, this benchmark is designed for evaluating the factuality capabilities of LLMs in Urdu (see \tabref{tab-urdufactqa}).

\begin{table}[t!]
\small
\centering
\begin{tabular}{L{2.8cm}|R{0.9cm}R{0.9cm}R{0.9cm}}
\toprule
\textbf{Dataset} & \textbf{\#True} & \textbf{\#False}  & \textbf{Total} \\
\midrule
\factcheckbench & 472 & 159 & 631 \\
\factoolqa & 177 & 56  & 233 \\
\bingcheck & 100 & 42 & 142 \\
\midrule
\urdufactbench & 749 & 257 & 1006 \\
\bottomrule
\end{tabular}
\caption{Statistics of \urdufactbench.}
\label{tab-urdufactbench}
\end{table}

\begin{table}[t!]
\small
\centering
\begin{tabular}{L{5.4cm}|R{0.9cm}}
\toprule
\textbf{Dataset} & \textbf{Size} \\
\midrule
\simpleqa  & 4,326 \\
\freshqa  & 600 \\
\midrule
\urdufactqa  & 4926 \\
\bottomrule
\end{tabular}
\caption{Statistics of \urdufactqa.}
\label{tab-urdufactqa}
\end{table}

Together, these benchmarks fill a critical gap in Urdu NLP by providing the first resources for reproducible, benchmarked research on factuality in low-resource settings.

\section{Framework}
\label{sec:framework}

To address the challenge of evaluating factuality in Urdu free-form text, we present \urdufactcheck, a set of three end-to-end pipelines specifically tailored for the Urdu language. The base framework consists of four core agent modules: \claimprocessor, \querygenerator, \retriever, and \verifier, drawing on well-established automated fact-checking frameworks, as illustrated in \figref{fig:architecture}~\cite{li-etal-2025-loki, iqbal-etal-2024-openfactcheck, xie-etal-2025-fire, chern2023factool}.

\subsection{Prompt Engineering for Core Modules}

The \urdufactcheck framework adopts an agentic architecture, where each module functions as a specialized agent with a distinct role. Prompts are designed to handle Urdu’s linguistic and contextual challenges, ensuring coherent outputs across the pipeline.

The \claimprocessor (\textsc{CP}) decomposes text into atomic, check-worthy claims. The \querygenerator (\textsc{QG}) produces two query types per claim: (i) question-based queries that conceal the fact and (ii) direct claim-based queries to improve retrieval. The \retriever (\textsc{RTV}) uses the Google SERP API\footnote{\url{https://serper.dev}} without prompt engineering, while the \verifier (\textsc{VFR}) assigns factuality labels, provides reasoning, and suggests corrections. Each prompt includes two–three examples, with full templates in \appref{app:prompts}.

\subsection{Evidence Boosting}

A major challenge in automated fact-checking for Urdu is the limited availability of high-quality evidence in the language. To address this, \urdufactcheck implements a multi-strategy evidence retrieval approach, consisting of three distinct strategies, that dynamically adapts to the difficulty and resource needs of each claim.

\paragraph{Monolingual Retrieval:}

This is a straightforward approach in which, for every Urdu query $q_\text{ur}$, the system retrieves evidence $E_\text{ur}$ in Urdu. This method ensures language consistency and computational efficiency, but struggles to provide relevant results for niche or globally underrepresented topics due to the scarcity of reliable Urdu web content. As a result, the evidence retrieved can sometimes be insufficient or only loosely related to the original claim.

\paragraph{Translated Retrieval:}

This strategy seeks to overcome the limitations of monolingual retrieval by translating the Urdu query $q_\text{ur}$ into English $q_\text{en}$ and conducting the web search in English to obtain evidence $E_\text{en}$. The retrieved evidence is then translated back into Urdu, resulting in $E_\text{en-ur}$, to maintain consistency with downstream modules. While this translation-based approach significantly improves evidence recall and quality by leveraging abundant English online sources, it incurs higher computational overhead and introduces potential risks of semantic drift during back-translation.

\paragraph{Thresholded Translated Retrieval:}

This approach combines the efficiency of monolingual retrieval with the robustness of translation-based search using a dynamic fallback mechanism. We introduce a thresholded evidence retrieval function, $\mathcal{R}(q_\text{ur}, \tau)$, which first attempts direct Urdu retrieval. The sufficiency of evidence $E_\text{ur}$ is assessed by comparing its cardinality $|E_\text{ur}|$ to a predefined threshold $\tau$ which represents the minimum evidence count.

If $|E_\text{ur}| \geq \tau$, the system proceeds with $E_\text{ur}$ for factual verification. Otherwise, $q_\text{ur}$ is translated into English ($q_\text{en}$) and additional evidence $E_\text{en}$ is retrieved using English search. This evidence is then translated back into Urdu $E_\text{en-ur}$. In such cases, both $E_\text{ur}$ and $E_\text{en-ur}$ are combined for downstream verification.

\vspace{-0.4cm}
\begin{equation}
\mathcal{R}(q, \tau) =
\begin{cases}
    E_{\text{ur}}, & \text{if } |E_{\text{ur}}| \geq \tau \\
    E_{\text{ur}} \cup E_{\text{en-ur}}, & \text{otherwise}
\end{cases}
\label{eq:evidence-retrieval}
\end{equation}

As shown in Equation~\ref{eq:evidence-retrieval}, this adaptive approach allows the system to default to efficient monolingual retrieval while guaranteeing broader verification coverage when Urdu evidence is insufficient. In such cases, the system dynamically invokes translation-based retrieval and combines evidence from both languages to strengthen verification. This design ensures that retrieval quality does not degrade even in scenarios where Urdu web content is sparse.

To support these transitions, we engineered dedicated prompts for Urdu-to-English and English-to-Urdu translation. All translation is performed by an LLM agent, which preserves meaning and maintains consistency across the pipeline. This setup not only improves recall but also ensures that retrieved evidence is usable in downstream modules. Overall, the tiered retrieval framework enables \urdufactcheck to balance accuracy and cost, while directly addressing the central challenge of evidence scarcity in Urdu fact-checking.

\section{Experiments}

To evaluate \urdufactcheck and our benchmarks, we conducted three experiments: (i) analyzing the effect of evidence thresholding on retrieval and verification, (ii) benchmarking automated fact-checkers with \urdufactbench, and (iii) assessing LLM factuality with \urdufactqa. We also report API costs for proprietary LLMs, GPU rental for open-source models, search engine query expenses, and total fact-checking time. Open-source experiments ran on an NVIDIA RTX 6000 GPU (\$0.79/hour), with each SerpAPI query costing about \$0.00105.

\subsection{Threshold Tuning}

A key hyperparameter in the retrieval pipeline is the evidence threshold $\tau$, which specifies the minimum number of Urdu snippets required before falling back to translation-based retrieval. To study trade-offs between recall, accuracy, and efficiency, we vary $\tau \in \{1, 3, 5, 7, 9\}$ and evaluate performance on the \factcheckbench subset of \urdufactbench, recording verification accuracy and retrieval cost. This identifies the optimal threshold balancing recall and cost. All experiments use \gptfouromini as the backbone model, with temperature 0 and a 2500-token limit; other parameters remain default.

\begin{figure}[t!]
\centering
\includegraphics[width=1\linewidth]{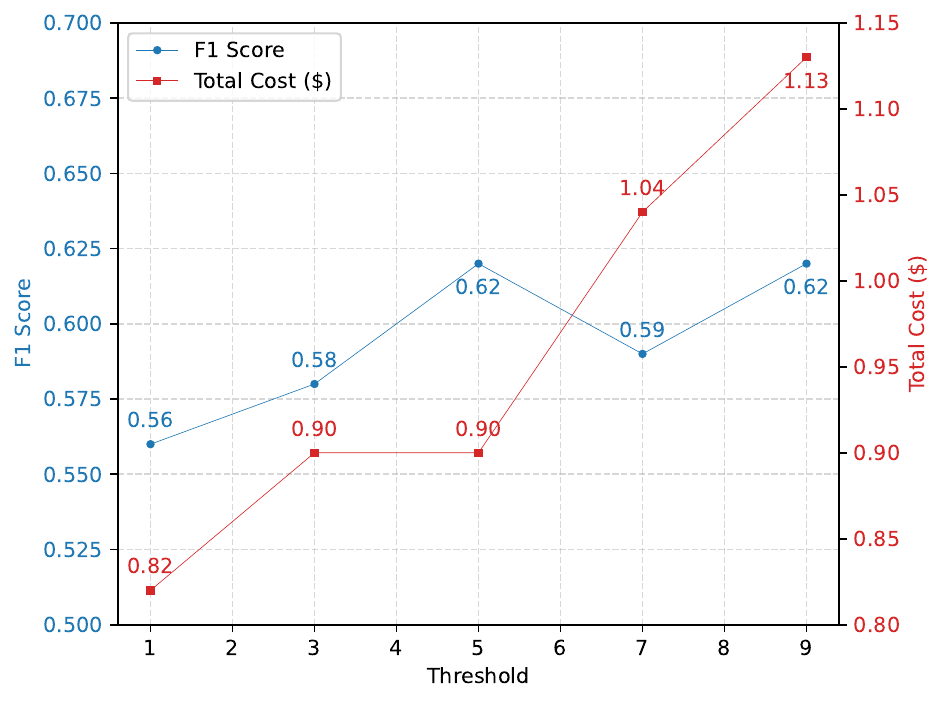}
\caption{Effect of evidence threshold $\tau$ on fact-checking performance and cost for the \factcheckbench subset of \urdufactbench. The blue line (left axis) shows the F1 score, while the red line (right axis) shows the total cost (\$). Higher thresholds increase cost but can improve F1 up to an optimal range before plateauing.}
\label{fig:threshholding}
\vspace{-0.6cm}
\end{figure}

\paragraph{Results Analysis}
Figure~\ref{fig:threshholding} shows the effect of varying the evidence threshold $\tau$ on both F1 score and total retrieval cost. As $\tau$ increases from 1 to 5, F1 score improves, but a dip is observed at $\tau = 7$ before rising again at $\tau = 9$. This pattern is not noise, but rather reflects that higher thresholds may introduce more loosely related or noisy snippets, which can hinder verification accuracy despite increasing the volume of evidence. The increase in cost is expected, as larger $\tau$ values trigger more frequent fallback to translation-based retrieval. 

Overall, setting $\tau$ in the range of 3–5 appears to provide a favorable balance between improved accuracy and manageable computational cost, making it a practical choice for deployment in resource-constrained or cost-sensitive scenarios. Based on these results, we use $\tau = 5$ as the default threshold for subsequent experiments, unless otherwise specified.

\subsection{LLM Selection for Fact-Checking}

Before conducting a full evaluation of \urdufactcheck, we first examined the performance of a set of LLMs on the \factcheckbench subset of \urdufactbench. The aim of this step was to identify which models provide the best balance between accuracy and cost, so that they could be used as backbone verifiers in later benchmarking experiments. We chose \factcheckbench for this initial study because it provides a diverse set of claims and has significant number of samples, making it a suitable testbed for comparing models under controlled conditions.

\begin{table*}[!ht]
\centering
\setlength{\tabcolsep}{6pt}
\begin{tabular}{@{}l|c|MMM|MMM@{}}
\toprule
 \multirow{2}{*}{\textbf{LLM}}  & \multirow{2}{*}{\textbf{\shortstack{LLM + Search\\ Cost (\$)}}} & \multicolumn{3}{c|}{\textbf{Label = True}} & \multicolumn{3}{c}{\textbf{Label = False}} \\ 
 & & Prec & Recall & F1 & Prec & Recall & F1 \\ 
 \midrule
\gptfourpointone                & 6.06+2.35 & 0.92 & 0.56 & 0.70 & 0.39 & 0.85 & 0.54 \\
\gptfourpointonemini            & 1.10+2.06 & 0.88 & 0.61 & 0.72 & 0.40 & 0.75 & 0.52 \\
\gptfouro                       & 7.42+2.32 & 0.90 & 0.56 & 0.69 & 0.38 & 0.80 & 0.52 \\
\gptfouromini                   & 0.35+1.87 & 0.92 & 0.48 & 0.63 & 0.36 & 0.87 & 0.51 \\
\claudesonnet                   & 21.6+2.66 & 0.90 & 0.44 & 0.59 & 0.34 & 0.85 & 0.49 \\
\claudehaiku                    & 5.71+2.73 & 0.85 & 0.40 & 0.54 & 0.30 & 0.79 & 0.44 \\
\midrule
\model{Mistral-Inst 7B}         & 1.84+1.22 & 0.80 & 0.39 & 0.52 & 0.30 & 0.62 & 0.40 \\
\model{Llama3.1-Inst 8B}        & 4.02+2.15 & 0.84 & 0.43 & 0.57 & 0.32 & 0.65 & 0.42 \\
\bottomrule
\end{tabular}
\caption{Fact-checking performance and cost of different language models on the \factcheckbench subset of \urdufactbench, using Thresholded Retrieval ($\tau = 5$). Results are reported separately for \textit{Label = True} and \textit{Label = False} in terms of precision, recall, and F1, alongside the combined LLM and search cost. Each cell is color-coded from red (lowest) to green (highest) within its column to highlight relative performance.}
\label{tab-languagemodels}
\vspace{0.5cm}
\setlength{\tabcolsep}{3pt}
\scriptsize
\centering
\renewcommand{\arraystretch}{1.2}
\resizebox{\textwidth}{!}{
\small
\begin{tabular}{@{}l|c|c|MMM|MMM|MMM|MMM|ccc@{}}
\toprule
\multicolumn{1}{c|}{\multirow{3}{*}{\textbf{Framework}}} & \multicolumn{1}{c|}{\multirow{3}{*} {\textbf{LLM}}}  & \multicolumn{1}{c|}{\multirow{3}{*} {\textbf{\shortstack{LLM + Search\\ Cost (\$)}}}} & \multicolumn{6}{c|}{\textbf{\urdufactbench - \factoolqa}} & \multicolumn{6}{c|}{\textbf{\urdufactbench - \bingcheck}} & \multicolumn{3}{c}{\textbf{Urdu}} \\
&   &   & \multicolumn{3}{c|}{\textbf{Label = True}} & \multicolumn{3}{c|}{\textbf{Label = False}}  & \multicolumn{3}{c|}{\textbf{Label = True}} & \multicolumn{3}{c|}{\textbf{Label = False}} & \multicolumn{3}{c}{\textbf{Language}} \\ 
& &  & Prec & Recall & F1 & Prec & Recall & F1 & Prec & Recall & F1 & Prec & Recall & F1 & CP & RTV & VFR\\
\midrule
Random                                  & -             & - & 0.58 & 0.77 & 0.66 & 0.46 & 0.26 & 0.33 & 0.58 & 0.77 & 0.66 & 0.46 & 0.26 & 0.33 & - & - & - \\
Always True                             & -             & - & 1.00 & 0.76 & 0.86 & 0.00 & 0.00 & 0.00 & 1.00 & 0.76 & 0.86 & 0.00 & 0.00 & 0.00 & - & - & - \\
Always False                            & -             & - & 0.00 & 0.00 & 0.00 & 1.00 & 0.24 & 0.39 & 0.00 & 0.00 & 0.00 & 1.00 & 0.24 & 0.39 & - & - & - \\
\midrule

\multirow{2}{*}{\factool}                   & \gptfouro     & 4.67+1.57  & 0.75 & 0.50 & 0.60 & 0.43 & 0.59 & 0.50 & 0.82 & 0.47 & 0.60 & 0.38 & 0.76 & 0.50  & $\times$ & $\times$ & $\times$ \\
                                            & \gptfouromini & 0.21+1.22  & 0.72 & 0.48 & 0.56 & 0.41 & 0.61 & 0.49 & 0.84 & 0.46 & 0.59 & 0.39 & 0.79 & 0.52  & $\times$ & $\times$ & $\times$ \\ 
\midrule \arrayrulecolor{white}
\multirow{2}{*}{\urdufactcheck}             & \gptfouro     & 4.87+1.61  & 0.84 & 0.63 & 0.72 & 0.35 & 0.63 & 0.45 & 0.87 & 0.41 & 0.56 & 0.39 & 0.86 & 0.54  & \checkmark & \checkmark & \checkmark \\
                                            & \gptfouromini & 0.22+1.24  & 0.87 & 0.53 & 0.65 & 0.33 & 0.75 & 0.46 & 0.87 & 0.45 & 0.59 & 0.34 & 0.84 & 0.48  & \checkmark & \checkmark & \checkmark \\  \midrule
\multirow{2}{*}{\urdufactcheck{ TH-TR-3}}   & \gptfouro     & 5.02+1.72  & 0.84 & 0.62 & 0.71 & 0.35 & 0.64 & 0.45 & 0.88 & 0.41 & 0.56 & 0.40 & 0.85 & 0.54 &  \checkmark & \checkmark & \checkmark \\
                                            & \gptfouromini & 0.24+1.37  & 0.83 & 0.48 & 0.61 & 0.29 & 0.68 & 0.41 & 0.87 & 0.47 & 0.61 & 0.40 & 0.84 & 0.55  & \checkmark & \checkmark & \checkmark \\ \midrule
\multirow{2}{*}{\urdufactcheck{ TH-TR-5}}   & \gptfouro     & 5.45+2.19  & 0.83 & 0.65 & 0.73 & 0.34 & 0.57 & 0.43 & 0.83 & 0.41 & 0.55 & 0.38 & 0.81 & 0.52  & \checkmark & \checkmark & \checkmark \\
                                            & \gptfouromini & 0.24+1.37  & 0.87 & 0.50 & 0.64 & 0.33 & 0.77 & 0.46 & 0.93 & 0.50 & 0.65 & 0.44 & 0.91 & 0.59  & \checkmark & \checkmark & \checkmark \\ \midrule
\multirow{2}{*}{\urdufactcheck{ TH-TR-7}}   & \gptfouro     & 5.20+2.38  & 0.84 & 0.67 & 0.75 & 0.35 & 0.59 & 0.44 & 0.80 & 0.40 & 0.53 & 0.35 & 0.77 & 0.49  & \checkmark & \checkmark & \checkmark \\
                                            & \gptfouromini & 0.28+1.59  & 0.87 & 0.53 & 0.66 & 0.34 & 0.79 & 0.48 & 0.89 & 0.48 & 0.62 & 0.42 & 0.86 & 0.56  & \checkmark & \checkmark & \checkmark \\ \midrule
\multirow{2}{*}{\urdufactcheck{ TH-TR-9}}   & \gptfouro     & 6.12+2.67  & 0.87 & 0.53 & 0.66 & 0.34 & 0.79 & 0.48 & 0.80 & 0.41 & 0.54 & 0.36 & 0.77 & 0.49  & \checkmark & \checkmark & \checkmark \\
                                            & \gptfouromini & 0.30+1.66  & 0.85 & 0.53 & 0.66 & 0.33 & 0.71 & 0.45 & 0.90 & 0.53 & 0.67 & 0.44 & 0.86 & 0.58  & \checkmark & \checkmark & \checkmark \\ \midrule
\multirow{2}{*}{\urdufactcheck{ TR}}        & \gptfouro     & 8.87+2.23  & 0.90 & 0.70 & 0.79 & 0.44 & 0.75 & 0.56 & 0.79 & 0.55 & 0.65 & 0.39 & 0.67 & 0.50  & \checkmark & \checkmark & \checkmark \\
                                            & \gptfouromini & 0.46+1.38  & 0.88 & 0.58 & 0.70 & 0.37 & 0.78 & 0.50 & 0.92 & 0.55 & 0.69 & 0.45 & 0.88 & 0.60 & \checkmark & \checkmark & \checkmark \\ 
\arrayrulecolor{black}
\bottomrule
\end{tabular}
}
\caption{Comparison of fact-checking performance across frameworks on \urdufactbench subsets (\factoolqa and \bingcheck). Each cell is color-coded from red (lowest) to green (highest) within its column to highlight relative performance. Metrics are reported separately for \textit{Label = True} and \textit{Label = False}, along with precision (Prec), recall, and F1. Cost values denote the combined expense of the LLM and search.}
\label{tab-verification}
\vspace{-1em}
\end{table*}

The models included in this evaluation fall into two groups. The proprietary group consisted of the GPT series~\citep{gpt-4o, gpt-o1} and the Claude series~\citep{claude3}, while the open-source group included \model{Mistral-Inst 7B}~\citep{mistral-7b} and \model{Llama3.1-Inst 8B}~\citep{dubey2024llama}. This selection allowed us to compare high-resource proprietary models with widely used open-source alternatives that are more accessible but often trained with smaller scale resources.

\paragraph{Results Analysis:}

The results in Table~\ref{tab-languagemodels} show that proprietary models generally outperform open-source ones on precision and recall for true and false labels. \gptfourpointone\ achieved the strongest overall performance, confirming its capacity for accurate factual verification in Urdu. However, smaller proprietary variants such as \gptfourpointonemini\ and \gptfouromini\ also performed competitively, reaching similar F1 scores at a fraction of the computational cost. In contrast, the open-source models \model{Mistral-Inst 7B} and \model{Llama3.1-Inst 8B} showed weaker performance, particularly on false labels, which suggests limitations in their ability to reliably detect incorrect claims in Urdu.

\begin{figure*}[!ht]
	\centering
	\includegraphics[width=1\linewidth]{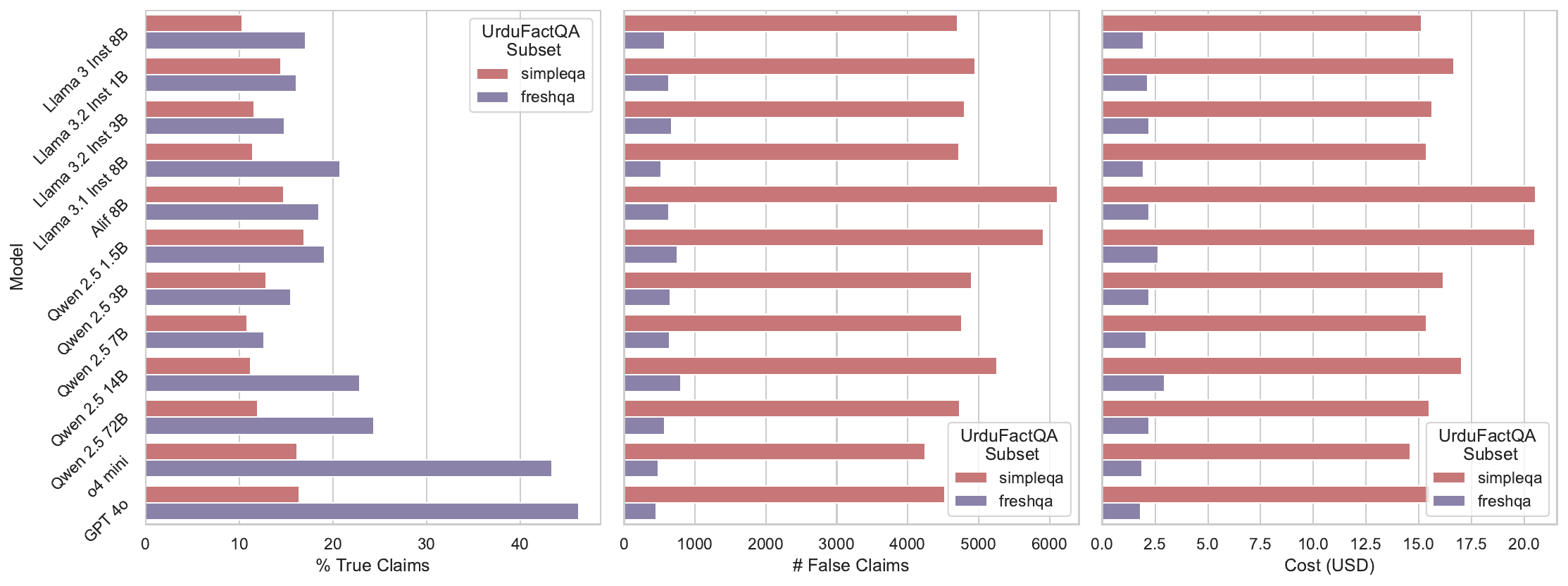}
	\caption{Automatic factuality evaluation results for 12 SOTA LLMS on \urdufactqa using \urdufactcheck{-TR}. \textit{left:} the percentage of true claims, \textit{center:} the number of false claims, and \textit{right:} the cost of using \urdufactcheck{-TR} in USD.}
	\label{fig:llm-evaluation}
\end{figure*}

A key observation from this experiment is the cost-performance trade-off. While large proprietary models provide the highest accuracy, smaller variants deliver comparable results at lower cost, making them practical for large-scale evaluation. Based on these findings, we selected \gptfouro\ and \gptfouromini\ as backbone models for subsequent experiments. These represent a complementary pair: one offering accuracy and the other efficiency. This ensured that benchmarking of \urdufactcheck\ was grounded in a balance of performance and efficiency for Urdu factual verification, while also providing a basis for future comparisons with other models.

\subsection{Fact-Checker Benchmarking}

To evaluate \urdufactcheck, we conducted experiments on two additional \urdufactbench subsets: \factoolqa and \bingcheck. As no end-to-end fact-checking systems exist for Urdu, direct comparisons are limited. We include \factool, but it produced unreliable outputs on Urdu text. For this reason, we did not extend comparisons to English-based fact-checkers, as their results would not provide a fair basis for Urdu verification, underscoring the need for Urdu-specific evaluation.
\pagebreak

Our evaluation included three variants of \urdufactcheck: (i) a monolingual retrieval pipeline, (ii) a fully translated retrieval pipeline (TR), and (iii) a thresholded translated retrieval pipeline (TH-TR). For the TH-TR variant, we varied the evidence threshold parameter $\tau$ across values 3, 5, 7, and 9, resulting in six distinct configurations of the system. All experiments were performed using two backbone language models: \gptfouro and \gptfouromini. This setup allowed us to study the trade-off between retrieval quality, accuracy, and cost under different system configurations.

\paragraph{Results Analysis:}

The results are reported in Table~\ref{tab-verification}. Across both subsets, all \urdufactcheck variants outperform \factool and trivial baselines. Among the three approaches, translation-based pipelines (TR and TH-TR) give the strongest results. For example, the TR variant with \gptfouro yields F1 scores up to 0.79 on true labels, showing the benefit of using English evidence to supplement limited Urdu web content. The TH-TR variants also achieve high accuracy while controlling cost, with $\tau=5$ providing a favorable balance between retrieval quality and efficiency.

In addition to accuracy, cost is an important factor in benchmarking fact-checking pipelines. Our analysis shows that \gptfouromini achieves performance close to \gptfouro but at a much lower cost, making it a practical choice for large-scale or resource-constrained applications. These findings demonstrate that while \urdufactcheck can be configured for higher accuracy through translation-based retrieval, lightweight variants offer a cost-effective alternative without a major loss in performance. This flexibility allows users to select configurations that best match their accuracy requirements and available resources.

\subsection{Evaluating LLM Factuality}

We evaluated the factual accuracy of twelve LLMs on the \urdufactqa benchmark. The evaluation included both proprietary and open-source models. The proprietary group consisted of \gptfouro and \model{o4-mini}. The open-source group included \model{Alif (8B)}, a Urdu specific model, as well as the \model{Llama-3-Inst (8B)}, \model{Llama-3.1-Inst (8B)}, \model{Llama-3.2-Inst (1B, 3B)}, and \model{Qwen 2.5 (1.5B, 3B, 7B, 14B, 72B)} families.

\begin{table*}[!ht]
\centering
\resizebox{\textwidth}{!}{\small
\begin{tabular}{l  p{13cm}  |C{0.7cm}C{0.7cm}C{0.7cm}|C{0.7cm}C{0.7cm}C{0.7cm}|C{0.7cm}C{0.7cm}C{0.7cm}|  C{1cm}}
\toprule
\multirow{3}{*}{\textbf{\textbf{Major Issue}}} & \multirow{3}{*}{\textbf{Error Type Description}}\ & \multicolumn{3}{c|}{\multirow{2}{*}{\textbf{\textbf{\factcheckbench}}}} & \multicolumn{3}{c|}{\multirow{2}{*}{\textbf{\textbf{\factool}}}} & \multicolumn{3}{c|}{\multirow{2}{*}{\textbf{\textbf{\bingcheck}}}} & \multirow{3}{*}{\textbf{\textbf{\#Total}}} \\
&&&&&&&&&&\\
& & \textbf{I} & \textbf{II} & \textbf{III} & \textbf{I} & \textbf{II} & \textbf{III} & \textbf{I} & \textbf{II} & \textbf{III} \\
    \midrule 
    \multirow{10}{3cm}{A. Dataset Issue} &
    \textbf{1. Invalid claim:} When the input consists only of a single entity such as a person's name or a place name or the claim is simply not check-worthy. For Example: \newline\null\hfill \foreignlanguage{urdu}{وہ چائے پی رہا تھا۔} or \foreignlanguage{urdu}{نیو یارک} &
    0 & 0 & 0 & 0 & 0 & 0 & 0 & 0 & 0 & 0\\
    & \textbf{2. Unclear, ambiguous, or subjective claim:} When the statement is vague, open to multiple interpretations, or not objectively verifiable. For Example: \newline \null\hfill\foreignlanguage{urdu}{انہوں نے میدان مار لیا۔} & 
    0 & 1 & 0 & 0 & 2 & 2 & 1 & 2 & 1 & 9\\
    & \textbf{3. False gold labels:} Cases where the original annotated label is incorrect. For Example the following statement is erroneously gold labeled as true: \newline \null\hfill \foreignlanguage{urdu}{سبز چائے پینے سے ذیابیطس مکمل طور پر ختم ہو جاتی ہے۔} & 
    0 & 0 & 0 & 0 & 0 & 0 & 0 & 0 & 0 & 0\\
    \midrule
    \multirow{13}{3cm}{B. Knowledge Issue} &
    \textbf{4. Requires complex scientific or domain-expert knowledge:} Cases where specialized expertise is necessary to verify the claim. For Example: \newline \null\hfill \foreignlanguage{urdu}{دعویٰ کہ بلیک ہول کے قریب وقت کی رفتار کم ہو جاتی ہے۔}  & 
    2 & 4 & 6 & 1 & 2 & 3 & 0 & 0 & 2 & 20\\
    & \textbf{5. Inaccurate parametric knowledge:} Cases where LLM-based verifiers make incorrect judgments because of erroneous or outdated knowledge stored in their parameters. For Example: \newline \null\hfill \foreignlanguage{urdu}{دعویٰ کہ پاکستان کا دارالحکومت کراچی ہے۔} & 
    4 & 8 & 5 & 4 & 2 & 5 & 0 & 0 & 3 & 31\\
    & \textbf{6. Insufficient or inaccurate externally collected knowledge (evidence):} Cases where verification fails due to (i) no external evidence retrieved and the model relies only on its own reasoning; (ii) evidence collected is incomplete and does not cover all aspects of a complex claim; (iii) evidence itself is inaccurate or contains errors. For Example: \newline \null\hfill \foreignlanguage{urdu}{دعویٰ کہ قراقرم میں دنیا کی پانچ بلند ترین چوٹیاں ہیں۔} & 
    11 & 5 & 1 & 6 & 6 & 2 & 12 & 9 & 7 & 59\\
    \midrule
    \multirow{7}{3cm}{C. LLM Reasoning} &
    \textbf{7. Incorrect reasoning:} Cases where the model draws invalid conclusions, either by dismissing relevant information from the claim or by introducing unsupported details.  & 
    3 & 2 & 6 & 9 & 8 & 6 & 4 & 5 & 6 & 49\\
    & \textbf{8. Strict reasoning:} Cases where the model relies too rigidly on a single source of knowledge: (i) strictly depending on collected evidence without using commonsense leads to wrong verification, whereas combining both could yield the correct result; (ii) overly strict reasoning based on parametric knowledge. For Example: \newline \null\hfill \foreignlanguage{urdu}{دعویٰ کہ فلُو اور اِنفلُوئنزا دو مختلف بیماریاں ہیں۔}). & 
    0 & 0 & 0 & 0 & 0 & 2 & 2 & 3 & 1 & 8\\
    \midrule
    \multirow{3}{3cm}{D. Debatable Opinion} &
    \textbf{9. Debatable opinions:} Cases involving controversial or disputed topics where multiple perspectives exist and no single view is universally accepted. For Example: \newline \null\hfill \foreignlanguage{urdu}{دعویٰ کہ موہنجو داڑو کی تہذیب آریاؤں نے قائم کی تھی}) & 
    0 & 0 & 2 & 0 & 0 & 0 & 1 & 1 & 0 & 4\\
     \midrule
    \textbf{Total} &  & 20 & 20 & 20 & 20 & 20 & 20 & 20 & 20 & 20 & 180\\ 
    \bottomrule
\end{tabular}
}
\caption{\textbf{Datasets error distribution, grouped into nine fine-grained types under four major issues.}. "I" represents the base \urdufactcheck pipeline, "II" refers to the version with translation-augmented retrieval, and "III" denotes the version with threshold-based rerouting.}
\label{tab:erroranalysis}
\end{table*}

For each question in \urdufactqa, responses were generated from all twelve models. Proprietary models used default OpenAI API parameters, while open-source models followed Hugging Face defaults. All outputs were automatically evaluated for factuality using the translation-based \urdufactcheck pipeline (TR variant) with \gptfouromini as verifier and Google Serper for retrieval. Prompts and templates from \urdufactqa were applied consistently.

\paragraph{Results Analysis:}

The results are shown in Figure~\ref{fig:llm-evaluation}. Proprietary models such as \gptfouro and \model{o4-mini} produced the highest percentage of factually correct responses across both subsets of \urdufactqa. In these models, the percentage of true claims reached up to 46\%. Open-source models, including \model{Alif-8B}, the \model{Llama-3} series, and the \model{Qwen} series, showed lower factual accuracy, often below 25\% true claims. Within the open-source group, \model{Qwen 2.5 (72B)} was the strongest model, though still behind proprietary systems.

The two subsets of \urdufactqa also showed different levels of difficulty. The \simpleqa subset contained a larger number of questions (4,326) and produced lower percentages of true claims, reflecting its greater challenge for LLMs. The \freshqa subset, with fewer questions (600), resulted in a higher proportion of correct answers and fewer false claims overall. This difference highlights the importance of dataset size and question design in assessing model factuality. Computational costs were comparable across models, with differences mainly due to model size rather than evaluation procedure.

In summary, the evaluation shows that proprietary models currently provide better factual accuracy for Urdu question answering, while open-source models have more limited performance. These findings confirm the value of large instruction-tuned models for factuality tasks in low-resource languages, and they emphasize the need for further development of competitive open-source models for Urdu. 

\subsection{Error Analysis}
Table~\ref{tab:erroranalysis} shows the distribution of nine error types, based on 20 randomly sampled errors from each evaluation. Most errors stem from \textbf{Knowledge Issues}, particularly insufficient or inaccurate external evidence (59 cases) and inaccurate parametric knowledge (31 cases). \textbf{LLM Reasoning} errors (49 incorrect reasoning, 8 strict reasoning) also reveal that models often dismiss relevant information, hallucinate unsupported details, or apply overly rigid reasoning strategies. 

Importantly, \urdufactcheck's translation-augmented and thresholded translation-augmented retrieval significantly reduce errors caused by low-quality evidence, confirming the effectiveness of multilingual strategies for Urdu fact-checking. While \textbf{Dataset Issues} and \textbf{Debatable Opinions} appear less often, their inclusion ensures coverage of ambiguity and contested claims. 

These findings demonstrate the complementary value of our two benchmarks: \urdufactbench enables fine-grained study of verification errors, while \urdufactqa highlights how factuality challenges surface in generative QA. Together, they provide a comprehensive basis for advancing factuality research in Urdu and related low-resource languages.
\section{Conclusion}

We introduced two new benchmarks for factuality evaluation in Urdu: \urdufactbench for claim verification and \urdufactqa for factual question answering. Built using translation and dual annotation, these are the first publicly available datasets enabling systematic study of factuality in Urdu, and filling a major resource gap for low-resource languages.

In order to demonstrate the utility of these new datasets, we further developed \urdufactcheck, a modular framework that integrates claim processing, evidence retrieval, and verification. It serves as a testbed for benchmarking fact-checking strategies and LLMs in Urdu. Using these resources, we conducted extensive experiments, including evaluations of twelve LLMs, revealing clear differences between proprietary and open-source models and underscoring both challenges and opportunities in Urdu factuality.

Our main contribution is the release of two high-quality benchmarks, supported by an open framework for experimentation. We hope that these resources will provide a foundation for advancing factuality assessment in low-resource languages, will foster cross-lingual transfer of methods, and will offer practical tools to address misinformation in Urdu and beyond.

\section*{Limitations and Future Work}

While \urdufactcheck represents a major step in factuality for Urdu, several limitations remain:

\paragraph{Evaluation Datasets}
\urdufactcheck relies heavily on the quality and diversity of the evaluation datasets. Although we have incorporated multiple benchmarks to ensure broad domain coverage, inherent biases and coverage gaps persist. Certain specialized domains may be underrepresented, potentially limiting the system’s robustness and generalizability for all types of factual claims.

\paragraph{Latency and Cost}
Automatic fact-checking with \urdufactcheck can incur substantial computational costs and latency, particularly when leveraging high-accuracy models and multi-stage retrieval strategies. These resource requirements may pose challenges for real-time applications or users with budgetary constraints.

\paragraph{Quality of Machine Translation}
The framework relies on machine translation when retrieving and processing evidence across Urdu and English. Despite careful prompt engineering and post-editing, translation errors can introduce semantic drift, loss of nuance, or context misinterpretation, potentially affecting both evidence quality and factuality judgments.

\paragraph{Handling Ambiguity and Subjectivity}
Some claims and questions may be inherently ambiguous, subjective, or context-dependent. The current framework is not equipped to distinguish between subjective assertions, nuanced opinions, or multi-faceted claims, which may impact the accuracy of factuality judgments in such cases.

\paragraph{Temporal Limitations}
\urdufactcheck does not explicitly model the temporal dynamics of factuality. As facts may change over time, especially in rapidly evolving domains, this can lead to mismatches between system judgments and the present state of knowledge. We are actively working on methods to integrate temporal awareness into future versions of the framework.

\paragraph{Dependence on External Knowledge Sources}
The framework’s reliance on external knowledge bases and web search engines introduces variability in the availability, reliability, and timeliness of evidence. Since web content is dynamic and not always up to date, the factual accuracy of retrieved information cannot be guaranteed in all scenarios.

\paragraph{Limited Human Evaluation}
While we perform automated evaluation of LLM outputs using \urdufactcheck, comprehensive human annotation and double-checking across all benchmarks is limited by human annotator availability and budget constraints. Automated metrics may not always fully capture nuanced or context-dependent factual errors that human experts could identify.

\section*{Ethical Statement}

The development and deployment of \urdufactcheck are guided by ethical principles to ensure responsible use and positive societal impact:

\paragraph{Transparency and Accountability}
We prioritize transparency by making our code, data, and evaluation protocols publicly available. This enables independent scrutiny and fosters community trust. We invite users and researchers to report issues and biases, promoting continual improvement of the framework.

\paragraph{Bias Mitigation}
We acknowledge the existence of potential biases in both language models and evaluation datasets. By integrating diverse benchmarks and supporting research into fair fact-checking, we aim to minimize the influence of bias on factuality assessments.

\paragraph{Social Impact}
Improving the factual accuracy of LLM outputs is central to combating misinformation and supporting informed public discourse. We believe \urdufactcheck can contribute meaningfully to these goals, especially in low-resource linguistic communities.

\bibliography{ref}

\begin{thebibliography}{43}
\providecommand{\natexlab}[1]{#1}

\bibitem[{Allcott and Gentzkow(2017)}]{allcott2017social}
Hunt Allcott and Matthew Gentzkow. 2017.
\newblock \href {https://www.aeaweb.org/articles?id=10.1257%2Fjep.31.2.211&ref=aurelius-ghost.ykaqi5.easypanel.host} {Social media and fake news in the 2016 election}.
\newblock \emph{Journal of Economic Perspectives}, 31(2):211--236.

\bibitem[{Amjad et~al.(2022)Amjad, Butt, Amjad, Zhila, Sidorov, and Gelbukh}]{amjad2022overview}
Maaz Amjad, Sabur Butt, Hamza~Imam Amjad, Alisa Zhila, Grigori Sidorov, and Alexander Gelbukh. 2022.
\newblock \href {https://arxiv.org/abs/2207.05133} {Overview of the shared task on fake news detection in {U}rdu at {FIRE} 2021}.
\newblock \emph{ArXiv preprint}, abs/2207.05133.

\bibitem[{Anthropic(2024{\natexlab{a}})}]{claude3}
Anthropic. 2024{\natexlab{a}}.
\newblock \href {https://assets.anthropic.com/m/61e7d27f8c8f5919/original/Claude-3-Model-Card.pdf} {The {C}laude 3 model family: {O}pus, {S}onnet, {H}aiku}.
\newblock \emph{Anthropic}.

\bibitem[{Anthropic(2024{\natexlab{b}})}]{claude35}
Anthropic. 2024{\natexlab{b}}.
\newblock \href {https://www-cdn.anthropic.com/fed9cc193a14b84131812372d8d5857f8f304c52/Model_Card_Claude_3_Addendum.pdf} {{C}laude 3.5 sonnet model card addendum}.
\newblock \emph{Anthropic}.

\bibitem[{Arechar et~al.(2023)Arechar, Allen, Berinsky, Cole, Epstein, Garimella, Gully, Lu, Ross, Stagnaro et~al.}]{arechar2023understanding}
Antonio~A Arechar, Jennifer Allen, Adam~J Berinsky, Rocky Cole, Ziv Epstein, Kiran Garimella, Andrew Gully, Jackson~G Lu, Robert~M Ross, Michael~N Stagnaro, and 1 others. 2023.
\newblock \href {https://www.nature.com/articles/s41562-023-01641-6} {Understanding and combatting misinformation across 16 countries on six continents}.
\newblock \emph{Nature Human Behaviour}, 7(9):1502--1513.

\bibitem[{Arif et~al.(2024)Arif, Farid, Athar, and Raza}]{arif2024uqa}
Samee Arif, Sualeha Farid, Awais Athar, and Agha~Ali Raza. 2024.
\newblock \href {https://aclanthology.org/2024.lrec-main.1497} {{UQA}: Corpus for {U}rdu question answering}.
\newblock In \emph{Proceedings of the 2024 Joint International Conference on Computational Linguistics, Language Resources and Evaluation}, LREC-COLING~'24, pages 17237--17244, Torino, Italia. ELRA and ICCL.

\bibitem[{Bang et~al.(2023)Bang, Cahyawijaya, Lee, Dai, Su, Wilie, Lovenia, Ji, Yu, Chung, Do, Xu, and Fung}]{bang2023multitask}
Yejin Bang, Samuel Cahyawijaya, Nayeon Lee, Wenliang Dai, Dan Su, Bryan Wilie, Holy Lovenia, Ziwei Ji, Tiezheng Yu, Willy Chung, Quyet~V. Do, Yan Xu, and Pascale Fung. 2023.
\newblock \href {https://doi.org/10.18653/v1/2023.ijcnlp-main.45} {A multitask, multilingual, multimodal evaluation of {C}hat{GPT} on reasoning, hallucination, and interactivity}.
\newblock In \emph{Proceedings of the 13th International Joint Conference on Natural Language Processing and the 3rd Conference of the Asia-Pacific Chapter of the Association for Computational Linguistics (Volume 1: Long Papers)}, IJCNLP-AACL~'23, pages 675--718, Nusa Dua, Bali. Association for Computational Linguistics.

\bibitem[{Borji(2023)}]{borji2023categorical}
Ali Borji. 2023.
\newblock \href {https://arxiv.org/abs/2302.03494} {A categorical archive of {ChatGPT} failures}.
\newblock \emph{ArXiv preprint}, abs/2302.03494.

\bibitem[{Chern et~al.(2023)Chern, Chern, Chen, Yuan, Feng, Zhou, He, Neubig, Liu et~al.}]{chern2023factool}
I~Chern, Steffi Chern, Shiqi Chen, Weizhe Yuan, Kehua Feng, Chunting Zhou, Junxian He, Graham Neubig, Pengfei Liu, and 1 others. 2023.
\newblock \href {https://arxiv.org/abs/2307.13528} {{FacTool}: Factuality detection in generative {AI}--{A} tool augmented framework for multi-task and multi-domain scenarios}.
\newblock \emph{ArXiv preprint}, abs/2307.13528.

\bibitem[{Chuang et~al.(2024)Chuang, Xie, Luo, Kim, Glass, and He}]{chuang2023dola}
Yung{-}Sung Chuang, Yujia Xie, Hongyin Luo, Yoon Kim, James~R. Glass, and Pengcheng He. 2024.
\newblock \href {https://openreview.net/forum?id=Th6NyL07na} {{DoLa}: Decoding by contrasting layers improves factuality in large language models}.
\newblock In \emph{Proceedings of the Twelfth International Conference on Learning Representations}, ICLR~'24, Vienna, Austria. OpenReview.net.

\bibitem[{Du et~al.(2021)Du, Dou, Xia, Cui, Ma, and Yu}]{du2021cross}
Jiangshu Du, Yingtong Dou, Congying Xia, Limeng Cui, Jing Ma, and Philip~S Yu. 2021.
\newblock \href {https://ieeexplore.ieee.org/abstract/document/9679918} {Cross-lingual {COVID-19} fake news detection}.
\newblock In \emph{Proceedings of the 2021 International Conference on Data Mining Workshops}, ICDMW~'21, pages 859--862. IEEE.

\bibitem[{Geng et~al.(2024)Geng, Cai, Wang, Koeppl, Nakov, and Gurevych}]{geng-etal-2024-survey}
Jiahui Geng, Fengyu Cai, Yuxia Wang, Heinz Koeppl, Preslav Nakov, and Iryna Gurevych. 2024.
\newblock \href {https://aclanthology.org/2024.naacl-long.366} {A survey of confidence estimation and calibration in large language models}.
\newblock In \emph{Proceedings of the 2024 Conference of the North American Chapter of the Association for Computational Linguistics: Human Language Technologies (Volume 1: Long Papers)}, NAACL~'24, pages 6577--6595, Mexico City, Mexico. Association for Computational Linguistics.

\bibitem[{Glava{\v{s}} et~al.(2020)Glava{\v{s}}, Karan, and Vuli{\'c}}]{glavavs2020xhate}
Goran Glava{\v{s}}, Mladen Karan, and Ivan Vuli{\'c}. 2020.
\newblock \href {https://doi.org/10.18653/v1/2020.coling-main.559} {{XH}ate-999: Analyzing and detecting abusive language across domains and languages}.
\newblock In \emph{Proceedings of the 28th International Conference on Computational Linguistics}, COLING~'20, pages 6350--6365, Barcelona, Spain (Online). International Committee on Computational Linguistics.

\bibitem[{Grattafiori et~al.(2024)Grattafiori, Dubey, Jauhri, Pandey, Kadian, Al-Dahle, Letman, Mathur, Schelten, Vaughan et~al.}]{dubey2024llama}
Aaron Grattafiori, Abhimanyu Dubey, Abhinav Jauhri, Abhinav Pandey, Abhishek Kadian, Ahmad Al-Dahle, Aiesha Letman, Akhil Mathur, Alan Schelten, Alex Vaughan, and 1 others. 2024.
\newblock \href {https://ai.meta.com/research/publications/the-llama-3-herd-of-models/} {The {L}lama 3 herd of models}.
\newblock \emph{ArXiv preprint}, abs/2407.21783.

\bibitem[{Guo et~al.(2022)Guo, Schlichtkrull, and Vlachos}]{guo2022survey}
Zhijiang Guo, Michael Schlichtkrull, and Andreas Vlachos. 2022.
\newblock \href {https://doi.org/10.1162/tacl_a_00454} {A survey on automated fact-checking}.
\newblock \emph{Transactions of the Association for Computational Linguistics}, 10:178--206.

\bibitem[{Gupta and Srikumar(2021)}]{gupta2021x}
Ashim Gupta and Vivek Srikumar. 2021.
\newblock \href {https://doi.org/10.18653/v1/2021.acl-short.86} {{X}-{F}act: A new benchmark dataset for multilingual fact checking}.
\newblock In \emph{Proceedings of the 59th Annual Meeting of the Association for Computational Linguistics and the 11th International Joint Conference on Natural Language Processing (Volume 2: Short Papers)}, IJCNLP~'21, pages 675--682, Online. Association for Computational Linguistics.

\bibitem[{Haider et~al.(2023)Haider, Luceri, Deb, Badawy, Peng, and Ferrara}]{haider2023detecting}
Samar Haider, Luca Luceri, Ashok Deb, Adam Badawy, Nanyun Peng, and Emilio Ferrara. 2023.
\newblock \href {https://dl.acm.org/doi/abs/10.1145/3543873.3587615} {Detecting social media manipulation in low-resource languages}.
\newblock In \emph{Companion Proceedings of the ACM Web Conference 2023}, WWW~'23, pages 1358--1364, Austin, TX, USA.

\bibitem[{Harris et~al.(2025)Harris, Liu, Hadi, Ahmad, and Alshara}]{harris2025benchmarking}
Sheetal Harris, Jinshuo Liu, Hassan~Jalil Hadi, Naveed Ahmad, and Mohammed~Ali Alshara. 2025.
\newblock \href {https://www.nature.com/articles/s41598-025-98271-x} {{Benchmarking Hook and Bait Urdu news dataset for domain-agnostic and multilingual fake news detection using large language models}}.
\newblock \emph{Scientific Reports}, 15(1):15553.

\bibitem[{Harris et~al.(2023)Harris, Liu, Hadi, and Cao}]{harris2023ax}
Sheetal Harris, Jinshuo Liu, Hassan~Jalil Hadi, and Yue Cao. 2023.
\newblock \href {https://ieeexplore.ieee.org/abstract/document/10538696} {{Ax-to-Grind Urdu}: {B}enchmark dataset for {U}rdu fake news detection}.
\newblock In \emph{Proceedings of the 2023 IEEE 22nd International Conference on Trust, Security and Privacy in Computing and Communications}, TrustCom~'23, pages 2440--2447. IEEE.

\bibitem[{Humprecht(2020)}]{humprecht2020they}
Edda Humprecht. 2020.
\newblock \href {https://www.tandfonline.com/doi/abs/10.1080/21670811.2019.1691031} {How do they debunk “fake news”? {A} cross-national comparison of transparency in fact checks}.
\newblock \emph{Digital journalism}, 8(3):310--327.

\bibitem[{Iqbal et~al.(2024)Iqbal, Wang, Wang, Georgiev, Geng, Gurevych, and Nakov}]{iqbal-etal-2024-openfactcheck}
Hasan Iqbal, Yuxia Wang, Minghan Wang, Georgi~Nenkov Georgiev, Jiahui Geng, Iryna Gurevych, and Preslav Nakov. 2024.
\newblock \href {https://doi.org/10.18653/v1/2024.emnlp-demo.23} {{OpenFactCheck}: A unified framework for factuality evaluation of {LLM}s}.
\newblock In \emph{Proceedings of the 2024 Conference on Empirical Methods in Natural Language Processing: System Demonstrations}, EMNLP~'24, pages 219--229, Miami, Florida, USA. Association for Computational Linguistics.

\bibitem[{Jiang et~al.(2023)Jiang, Sablayrolles, Mensch, Bamford, Chaplot, de~Las~Casas, Bressand, Lengyel, Lample, Saulnier, Lavaud, Lachaux, Stock, Scao, Lavril, Wang, Lacroix, and Sayed}]{mistral-7b}
Albert~Q. Jiang, Alexandre Sablayrolles, Arthur Mensch, Chris Bamford, Devendra~Singh Chaplot, Diego de~Las~Casas, Florian Bressand, Gianna Lengyel, Guillaume Lample, Lucile Saulnier, L{\'{e}}lio~Renard Lavaud, Marie{-}Anne Lachaux, Pierre Stock, Teven~Le Scao, Thibaut Lavril, Thomas Wang, Timoth{\'{e}}e Lacroix, and William~El Sayed. 2023.
\newblock \href {https://arxiv.org/abs/2310.06825} {Mistral 7b}.
\newblock \emph{ArXiv preprint}, abs/2310.06825.

\bibitem[{Kazi and Khoja(2021)}]{kazi2021uquad1}
Samreen Kazi and Shakeel Khoja. 2021.
\newblock \href {https://arxiv.org/abs/2111.01543} {Uquad1.0: Development of an {U}rdu question answering training data for machine reading comprehension}.
\newblock \emph{ArXiv preprint}, abs/2111.01543.

\bibitem[{Li et~al.(2025)Li, Han, Wang, Wang, Wang, Xing, Geng, Zhai, Nakov, and Baldwin}]{li-etal-2025-loki}
Haonan Li, Xudong Han, Hao Wang, Yuxia Wang, Minghan Wang, Rui Xing, Yilin Geng, Zenan Zhai, Preslav Nakov, and Timothy Baldwin. 2025.
\newblock \href {https://aclanthology.org/2025.coling-demos.4/} {Loki: {A}n open-source tool for fact verification}.
\newblock In \emph{Proceedings of the 31st International Conference on Computational Linguistics: System Demonstrations}, COLING~'25, pages 28--36, Abu Dhabi, UAE. Association for Computational Linguistics.

\bibitem[{Li et~al.(2024)Li, Peng, Galley, Gao, and Zhang}]{li2023self}
Miaoran Li, Baolin Peng, Michel Galley, Jianfeng Gao, and Zhu Zhang. 2024.
\newblock \href {https://aclanthology.org/2024.findings-naacl.12} {{S}elf-{C}hecker: {P}lug-and-{P}lay modules for fact-checking with large language models}.
\newblock In \emph{Findings of the Association for Computational Linguistics}, NAACL~'24, pages 163--181, Mexico City, Mexico. Association for Computational Linguistics.

\bibitem[{Min et~al.(2023)Min, Krishna, Lyu, Lewis, Yih, Koh, Iyyer, Zettlemoyer, and Hajishirzi}]{min2023factscore}
Sewon Min, Kalpesh Krishna, Xinxi Lyu, Mike Lewis, Wen-tau Yih, Pang Koh, Mohit Iyyer, Luke Zettlemoyer, and Hannaneh Hajishirzi. 2023.
\newblock \href {https://doi.org/10.18653/v1/2023.emnlp-main.741} {{FA}ct{S}core: {F}ine-grained atomic evaluation of factual precision in long form text generation}.
\newblock In \emph{Proceedings of the 2023 Conference on Empirical Methods in Natural Language Processing}, EMNLP~'23, pages 12076--12100, Singapore. Association for Computational Linguistics.

\bibitem[{OpenAI et~al.(2023)OpenAI, Achiam, Adler, Agarwal, Ahmad, Akkaya, Aleman, Almeida, Altenschmidt, Altman, Anadkat, Avila, Babuschkin, Balaji, Balcom, Baltescu, Bao, Bavarian, Belgum, Bello, Berdine, Bernadett-Shapiro, Berner, Bogdonoff, Boiko, Boyd, Brakman, Brockman, Brooks, Brundage, Button, Cai, Campbell, Cann, Carey, Carlson, Carmichael, Chan, Chang, Chantzis, Chen, Chen, Chen, Chen, Chen, Chess, Cho, Chu, Chung, Cummings, Currier, Dai, Decareaux, Degry, Deutsch, Deville, Dhar, Dohan, Dowling, Dunning, Ecoffet, Eleti, Eloundou, Farhi, Fedus, Felix, Fishman, Forte, Fulford, Gao, Georges, Gibson, Goel, Gogineni, Goh, Gontijo-Lopes, Gordon, Grafstein, Gray, Greene, Gross, Gu, Guo, Hallacy, Han, Harris, He, Heaton, Heidecke, Hesse, Hickey, Hickey, Hoeschele, Houghton, Hsu, Hu, Hu, Huizinga, Jain, Jain, Jang, Jiang, Jiang, Jin, Jin, Jomoto, Jonn, Jun, Kaftan, Łukasz Kaiser, Kamali, Kanitscheider, Keskar, Khan, Kilpatrick, Kim, Kim, Kim, Kirchner, Kiros, Knight, Kokotajlo, Łukasz Kondraciuk,
  Kondrich, Konstantinidis, Kosic, Krueger, Kuo, Lampe, Lan, Lee, Leike, Leung, Levy, Li, Lim, Lin, Lin, Litwin, Lopez, Lowe, Lue, Makanju, Malfacini, Manning, Markov, Markovski, Martin, Mayer, Mayne, McGrew, McKinney, McLeavey, McMillan, McNeil, Medina, Mehta, Menick, Metz, Mishchenko, Mishkin, Monaco, Morikawa, Mossing, Mu, Murati, Murk, Mély, Nair, Nakano, Nayak, Neelakantan, Ngo, Noh, Ouyang, O'Keefe, Pachocki, Paino, Palermo, Pantuliano, Parascandolo, Parish, Parparita, Passos, Pavlov, Peng, Perelman, de~Avila Belbute~Peres, Petrov, de~Oliveira~Pinto, Michael, Pokorny, Pokrass, Pong, Powell, Power, Power, Proehl, Puri, Radford, Rae, Ramesh, Raymond, Real, Rimbach, Ross, Rotsted, Roussez, Ryder, Saltarelli, Sanders, Santurkar, Sastry, Schmidt, Schnurr, Schulman, Selsam, Sheppard, Sherbakov, Shieh, Shoker, Shyam, Sidor, Sigler, Simens, Sitkin, Slama, Sohl, Sokolowsky, Song, Staudacher, Such, Summers, Sutskever, Tang, Tezak, Thompson, Tillet, Tootoonchian, Tseng, Tuggle, Turley, Tworek, Uribe, Vallone,
  Vijayvergiya, Voss, Wainwright, Wang, Wang, Wang, Ward, Wei, Weinmann, Welihinda, Welinder, Weng, Weng, Wiethoff, Willner, Winter, Wolrich, Wong, Workman, Wu, Wu, Wu, Xiao, Xu, Yoo, Yu, Yuan, Zaremba, Zellers, Zhang, Zhang, Zhao, Zheng, Zhuang, Zhuk, and Zoph}]{gpt-4o}
OpenAI, Josh Achiam, Steven Adler, Sandhini Agarwal, Lama Ahmad, Ilge Akkaya, Florencia~Leoni Aleman, Diogo Almeida, Janko Altenschmidt, Sam Altman, Shyamal Anadkat, Red Avila, Igor Babuschkin, Suchir Balaji, Valerie Balcom, Paul Baltescu, Haiming Bao, Mohammad Bavarian, Jeff Belgum, and 262 others. 2023.
\newblock \href {https://arxiv.org/abs/2303.08774} {{GPT}-4 technical report}.
\newblock \emph{ArXiv preprint}, abs/2303.08774.

\bibitem[{OpenAI et~al.(2024)OpenAI, Jaech, Kalai, Lerer, Richardson, El-Kishky, Low, Helyar, Madry, Beutel, Carney, Iftimie, Karpenko, Passos, Neitz, Prokofiev, Wei, Tam, Bennett, Kumar, Saraiva, Vallone, Duberstein, Kondrich, Mishchenko, Applebaum, Jiang, Nair, Zoph, Ghorbani, Rossen, Sokolowsky, Barak, McGrew, Minaiev, Hao, Baker, Houghton, McKinzie, Eastman, Lugaresi, Bassin, Hudson, Li, de~Bourcy, Voss, Shen, Zhang, Koch, Orsinger, Hesse, Fischer, Chan, Roberts, Kappler, Levy, Selsam, Dohan, Farhi, Mely, Robinson, Tsipras, Li, Oprica, Freeman, Zhang, Wong, Proehl, Cheung, Mitchell, Wallace, Ritter, Mays, Wang, Such, Raso, Leoni, Tsimpourlas, Song, von Lohmann, Sulit, Salmon, Parascandolo, Chabot, Zhao, Brockman, Leclerc, Salman, Bao, Sheng, Andrin, Bagherinezhad, Ren, Lightman, Chung, Kivlichan, O'Connell, Osband, Gilaberte, Akkaya, Kostrikov, Sutskever, Kofman, Pachocki, Lennon, Wei, Harb, Twore, Feng, Yu, Weng, Tang, Yu, Candela, Palermo, Parish, Heidecke, Hallman, Rizzo, Gordon, Uesato, Ward,
  Huizinga, Wang, Chen, Xiao, Singhal, Nguyen, Cobbe, Shi, Wood, Rimbach, Gu-Lemberg, Liu, Lu, Stone, Yu, Ahmad, Yang, Liu, Maksin, Ho, Fedus, Weng, Li, McCallum, Held, Kuhn, Kondraciuk, Kaiser, Metz, Boyd, Trebacz, Joglekar, Chen, Tintor, Meyer, Jones, Kaufer, Schwarzer, Shah, Yatbaz, Guan, Xu, Yan, Glaese, Chen, Lampe, Malek, Wang, Fradin, McClay, Pavlov, Wang, Wang, Murati, Bavarian, Rohaninejad, McAleese, Chowdhury, Chowdhury, Ryder, Tezak, Brown, Nachum, Boiko, Murk, Watkins, Chao, Ashbourne, Izmailov, Zhokhov, Dias, Arora, Lin, Lopes, Gaon, Miyara, Leike, Hwang, Garg, Brown, James, Shu, Cheu, Greene, Jain, Altman, Toizer, Toyer, Miserendino, Agarwal, Hernandez, Baker, McKinney, Yan, Zhao, Hu, Santurkar, Chaudhuri, Zhang, Fu, Papay, Lin, Balaji, Sanjeev, Sidor, Broda, Clark, Wang, Gordon, Sanders, Patwardhan, Sottiaux, Degry, Dimson, Zheng, Garipov, Stasi, Bansal, Creech, Peterson, Eloundou, Qi, Kosaraju, Monaco, Pong, Fomenko, Zheng, Zhou, McCabe, Zaremba, Dubois, Lu, Chen, Cha, Bai, He, Zhang, Wang,
  Shao, and Li}]{gpt-o1}
OpenAI, Aaron Jaech, Adam Kalai, Adam Lerer, Adam Richardson, Ahmed El-Kishky, Aiden Low, Alec Helyar, Aleksander Madry, Alex Beutel, Alex Carney, Alex Iftimie, Alex Karpenko, Alex~Tachard Passos, Alexander Neitz, Alexander Prokofiev, Alexander Wei, Allison Tam, Ally Bennett, and 243 others. 2024.
\newblock \href {https://arxiv.org/abs/2412.16720} {{O}pen{AI} o1 system card}.
\newblock \emph{ArXiv preprint}, abs/2412.16720.

\bibitem[{Organization et~al.(2023)}]{world2023overview}
World~Health Organization and 1 others. 2023.
\newblock \href {https://pesquisa.bvsalud.org/portal/resource/pt/who-370860} {An overview of infodemic management during {COVID-19} pandemic, january 2020--july 2022}.
\newblock \emph{An overview of infodemic management during {COVID-19} pandemic, January 2020--July 2022}.

\bibitem[{Pogue(2017)}]{pogue2017stamp}
David Pogue. 2017.
\newblock \href {https://dialnet.unirioja.es/servlet/articulo?codigo=6538385} {How to stamp out fake news}.
\newblock \emph{Scientific American}, 316(2):24--24.

\bibitem[{Raja et~al.(2023)Raja, Soni, and Borgohain}]{raja2023fake}
Eduri Raja, Badal Soni, and Samir~Kumar Borgohain. 2023.
\newblock \href {https://www.sciencedirect.com/science/article/abs/pii/S0952197623010618} {Fake news detection in {D}ravidian languages using transfer learning with adaptive finetuning}.
\newblock \emph{Engineering Applications of Artificial Intelligence}, 126:106877.

\bibitem[{Rajpurkar et~al.(2018)Rajpurkar, Jia, and Liang}]{rajpurkar-etal-2018-know}
Pranav Rajpurkar, Robin Jia, and Percy Liang. 2018.
\newblock \href {https://doi.org/10.18653/v1/P18-2124} {Know what you don{'}t know: Unanswerable questions for {SQ}u{AD}}.
\newblock In \emph{Proceedings of the 56th Annual Meeting of the Association for Computational Linguistics (Volume 2: Short Papers)}, ACL~'18, pages 784--789, Melbourne, Australia. Association for Computational Linguistics.

\bibitem[{Team et~al.(2024)Team, Georgiev, Lei, Burnell, Bai, Gulati, Tanzer, Vincent, Pan, Wang et~al.}]{team2024gemini}
Gemini Team, Petko Georgiev, Ving~Ian Lei, Ryan Burnell, Libin Bai, Anmol Gulati, Garrett Tanzer, Damien Vincent, Zhufeng Pan, Shibo Wang, and 1 others. 2024.
\newblock \href {https://arxiv.org/abs/2403.05530} {Gemini 1.5: Unlocking multimodal understanding across millions of tokens of context}.
\newblock \emph{ArXiv preprint}, abs/2403.05530.

\bibitem[{Tie et~al.(2024)Tie, Yao, Li, Ahmed, Wang, and Zhou}]{tie2024llms}
Jiessie Tie, Bingsheng Yao, Tianshi Li, Syed~Ishtiaque Ahmed, Dakuo Wang, and Shurui Zhou. 2024.
\newblock \href {https://arxiv.org/abs/2411.09916} {{LLMs} are imperfect, then what? {An} empirical study on {LLM} failures in software engineering}.
\newblock \emph{ArXiv preprint}, abs/2411.09916.

\bibitem[{Trilling et~al.(2017)Trilling, Tolochko, and Burscher}]{trilling2017newsworthiness}
Damian Trilling, Petro Tolochko, and Bj{\"o}rn Burscher. 2017.
\newblock \href {https://journals.sagepub.com/doi/abs/10.1177/1077699016654682} {From newsworthiness to shareworthiness: How to predict news sharing based on article characteristics}.
\newblock \emph{Journalism \& mass communication quarterly}, 94(1):38--60.

\bibitem[{Vosoughi et~al.(2018)Vosoughi, Roy, and Aral}]{vosoughi2018spread}
Soroush Vosoughi, Deb Roy, and Sinan Aral. 2018.
\newblock \href {https://www.science.org/doi/abs/10.1126/science.aap9559} {The spread of true and false news online}.
\newblock \emph{science}, 359(6380):1146--1151.

\bibitem[{Vu et~al.(2024)Vu, Iyyer, Wang, Constant, Wei, Wei, Tar, Sung, Zhou, Le, and Luong}]{vu-etal-2024-freshllms}
Tu~Vu, Mohit Iyyer, Xuezhi Wang, Noah Constant, Jerry Wei, Jason Wei, Chris Tar, Yun-Hsuan Sung, Denny Zhou, Quoc Le, and Thang Luong. 2024.
\newblock \href {https://doi.org/10.18653/v1/2024.findings-acl.813} {{F}resh{LLM}s: Refreshing large language models with search engine augmentation}.
\newblock In \emph{Findings of the Association for Computational Linguistics: ACL 2024}, pages 13697--13720, Bangkok, Thailand. Association for Computational Linguistics.

\bibitem[{Wang et~al.(2024{\natexlab{a}})Wang, Gangi~Reddy, Mujahid, Arora, Rubashevskii, Geng, Mohammed~Afzal, Pan, Borenstein, Pillai, Augenstein, Gurevych, and Nakov}]{wang-etal-2024-factcheck}
Yuxia Wang, Revanth Gangi~Reddy, Zain~Muhammad Mujahid, Arnav Arora, Aleksandr Rubashevskii, Jiahui Geng, Osama Mohammed~Afzal, Liangming Pan, Nadav Borenstein, Aditya Pillai, Isabelle Augenstein, Iryna Gurevych, and Preslav Nakov. 2024{\natexlab{a}}.
\newblock \href {https://doi.org/10.18653/v1/2024.findings-emnlp.830} {Factcheck-bench: Fine-grained evaluation benchmark for automatic fact-checkers}.
\newblock In \emph{Findings of the Association for Computational Linguistics}, EMNLP~'24, pages 14199--14230, Miami, Florida, USA. Association for Computational Linguistics.

\bibitem[{Wang et~al.(2025)Wang, Wang, Iqbal, Georgiev, Geng, Gurevych, and Nakov}]{wang-etal-2025-openfactcheck}
Yuxia Wang, Minghan Wang, Hasan Iqbal, Georgi~N. Georgiev, Jiahui Geng, Iryna Gurevych, and Preslav Nakov. 2025.
\newblock \href {https://aclanthology.org/2025.coling-main.755/} {{O}pen{F}act{C}heck: Building, benchmarking customized fact-checking systems and evaluating the factuality of claims and {LLM}s}.
\newblock In \emph{Proceedings of the 31st International Conference on Computational Linguistics}, COLING~'25, pages 11399--11421, Abu Dhabi, UAE. Association for Computational Linguistics.

\bibitem[{Wang et~al.(2024{\natexlab{b}})Wang, Wang, Manzoor, Liu, Georgiev, Das, and Nakov}]{wang-etal-2024-factuality}
Yuxia Wang, Minghan Wang, Muhammad~Arslan Manzoor, Fei Liu, Georgi~Nenkov Georgiev, Rocktim~Jyoti Das, and Preslav Nakov. 2024{\natexlab{b}}.
\newblock \href {https://doi.org/10.18653/v1/2024.emnlp-main.1088} {Factuality of large language models: A survey}.
\newblock In \emph{Proceedings of the 2024 Conference on Empirical Methods in Natural Language Processing}, EMNLP~'24, pages 19519--19529, Miami, Florida, USA. Association for Computational Linguistics.

\bibitem[{Wei et~al.(2024)Wei, Karina, Chung, Jiao, Papay, Glaese, Schulman, and Fedus}]{wei2024measuring}
Jason Wei, Nguyen Karina, Hyung~Won Chung, Yunxin~Joy Jiao, Spencer Papay, Amelia Glaese, John Schulman, and William Fedus. 2024.
\newblock \href {https://arxiv.org/abs/2411.04368} {Measuring short-form factuality in large language models}.
\newblock \emph{ArXiv preprint}, abs/2411.04368.

\bibitem[{Xie et~al.(2025)Xie, Xing, Wang, Geng, Iqbal, Sahnan, Gurevych, and Nakov}]{xie-etal-2025-fire}
Zhuohan Xie, Rui Xing, Yuxia Wang, Jiahui Geng, Hasan Iqbal, Dhruv Sahnan, Iryna Gurevych, and Preslav Nakov. 2025.
\newblock \href {https://doi.org/10.18653/v1/2025.findings-naacl.158} {{FIRE}: Fact-checking with iterative retrieval and verification}.
\newblock In \emph{Findings of the Association for Computational Linguistics}, NAACL~'25, pages 2901--2914, Albuquerque, New Mexico. Association for Computational Linguistics.

\bibitem[{Zimmer et~al.(2019)Zimmer, Scheibe, Stock, and Stock}]{zimmer2019fake}
Franziska Zimmer, Katrin Scheibe, Mechtild Stock, and Wolfgang~G Stock. 2019.
\newblock \href {https://repository.kisti.re.kr/handle/10580/13511} {Fake news in social media: Bad algorithms or biased users?}
\newblock \emph{Journal of Information Science Theory and Practice}, 7(2):40--53.

\end{thebibliography}

\appendix

\section{Pre-Translation Prompt for Dataset Generation}
\label{app:pre-translation}

To speed up annotation for \urdufactqa and \urdufactbench, we designed a structured pre-translation prompt. It instructed the LLM to translate claim–label pairs into formal Urdu. The full prompt appears in \autoref{fig:translation-guidelines}.

\begin{codebox}
\renewcommand{\baselinestretch}{1.2} 
{\fontsize{8.5pt}{10.5pt}\selectfont
\ttfamily
You are an expert Urdu translator. Your task is to translate the following claim-label pairs from English to Urdu.\\

\#\#\# Instructions\\
- Translate both the \textit{claim} and \textit{label} into \textit{formal, fluent Urdu}.\\
- Use correct \textit{masculine/feminine grammatical forms} in Urdu.\\
- Translate \textit{proper nouns} only if a widely accepted Urdu version exists (e.g., "India" → \foreignlanguage{urdu}{بھارت}, "Syria" → \foreignlanguage{urdu}{شام}).\\
- Avoid translating proper nouns when they appear in the name of an organization.\\
- Retain \textit{technical or factual terms} (e.g., award names, organization names) in transliterated form, where appropriate.\\
- Translate \textit{dates} into proper Urdu format (e.g., "Jan 1, 2020" → "2020 \foreignlanguage{urdu}{یکم جنوری}").\\

\#\#\# Important Formatting Guidelines\\
1. \textit{English acronyms and abbreviations} (e.g., IEEE, NASA, UNESCO):\\
   - \textit{Do not} translate or transliterate.\\
   - Place them at a natural position in the Urdu sentence (ideally after the date or subject).\\
   - Avoid starting Urdu sentences with acronyms or left-to-right (LTR) text.\\

2. \textit{Western numerals and LTR elements} (e.g., 2022, 7.8.8, Notepad++):\\
   - \textit{Do not} convert numerals to Urdu words.\\
   - Always place an Urdu phrase before such elements to maintain proper right-to-left (RTL) sentence flow.\\
   - This applies to acronyms, version numbers, software/product names, etc.\\

   Incorrect (structurally broken):\\
   a. \hfill 2010 \foreignlanguage{urdu}{سال} IEEE \foreignlanguage{urdu}{فرینک روزن بلیٹ ایوارڈ کس کو دیا گیا؟}
   
   ~  \hfill \foreignlanguage{urdu}{میں}
   
   b. \hfill 2 of January of 2019
   
   c. \hfill \foreignlanguage{urdu}{رگبی یورپ چیمپئن شپ کا حصہ بننے والے اسپین اور رومانیہ}
   
   ~  \hfill \foreignlanguage{urdu}{کو اسپین کے لیے تمام} 2022 \foreignlanguage{urdu}{فروری} 27 \foreignlanguage{urdu}{کے درمیان رگبی میچ }
   
   \hfill 2022 \foreignlanguage{urdu}{کنورژنز کس کھلاڑی نے اسکور کیے؟}

   Correct (natural Urdu structure):\\
   a. \hfill \foreignlanguage{urdu}{کس کو دیا } \foreignlanguage{urdu}{میں} 2010\foreignlanguage{urdu}{فرینک روزن بلیٹ ایوارڈ سال} IEEE
   
   ~  \hfill \foreignlanguage{urdu}{گیا؟}
   
   b. \hfill \foreignlanguage{urdu}{جنوری کو} 2 \foreignlanguage{urdu}{میں} 2019 \foreignlanguage{urdu}{سال}

   c. \hfill \foreignlanguage{urdu}{کا حصہ بننے والے اسپین اور} 2022 \foreignlanguage{urdu}{رگبی یورپ چیمپئن شپ}

   ~ \hfill \hfill \foreignlanguage{urdu}{کو اسپین} 2022 \foreignlanguage{urdu}{فروری} 27 \foreignlanguage{urdu}{رومانیہ کے درمیان رگبی میچ میں}

   ~ \hfill \foreignlanguage{urdu}{اسپین کے لیے تمام کنورژنز کس کھلاڑی نے اسکور کیے؟}\\
   
3. Ensure the final Urdu sentence is:\\
   - Grammatically correct\\
   - Visually aligned for RTL display\\
   - Fluent and natural to read\\

\#\#\# Examples\\
Here are a few examples of claims and expected translations:\\
\{examples\}\\

\#\#\# Translation\\
claim: \{claim\}\\
label: \{label\}\\

\#\#\# Format Instructions:\\
\{format\_instructions\}
}
\end{codebox}
\captionof{lstlisting}{Pre-translation prompt used to guide LLM in generating Urdu translations of claims and labels. The prompt defines translation rules for proper nouns, acronyms, numerals, and dates, ensuring consistent and fluent output across the dataset.}
\label{fig:translation-guidelines}

\vspace{2cm}

\begin{figure*}[!t]
  \centering
  \includegraphics[width=\textwidth]{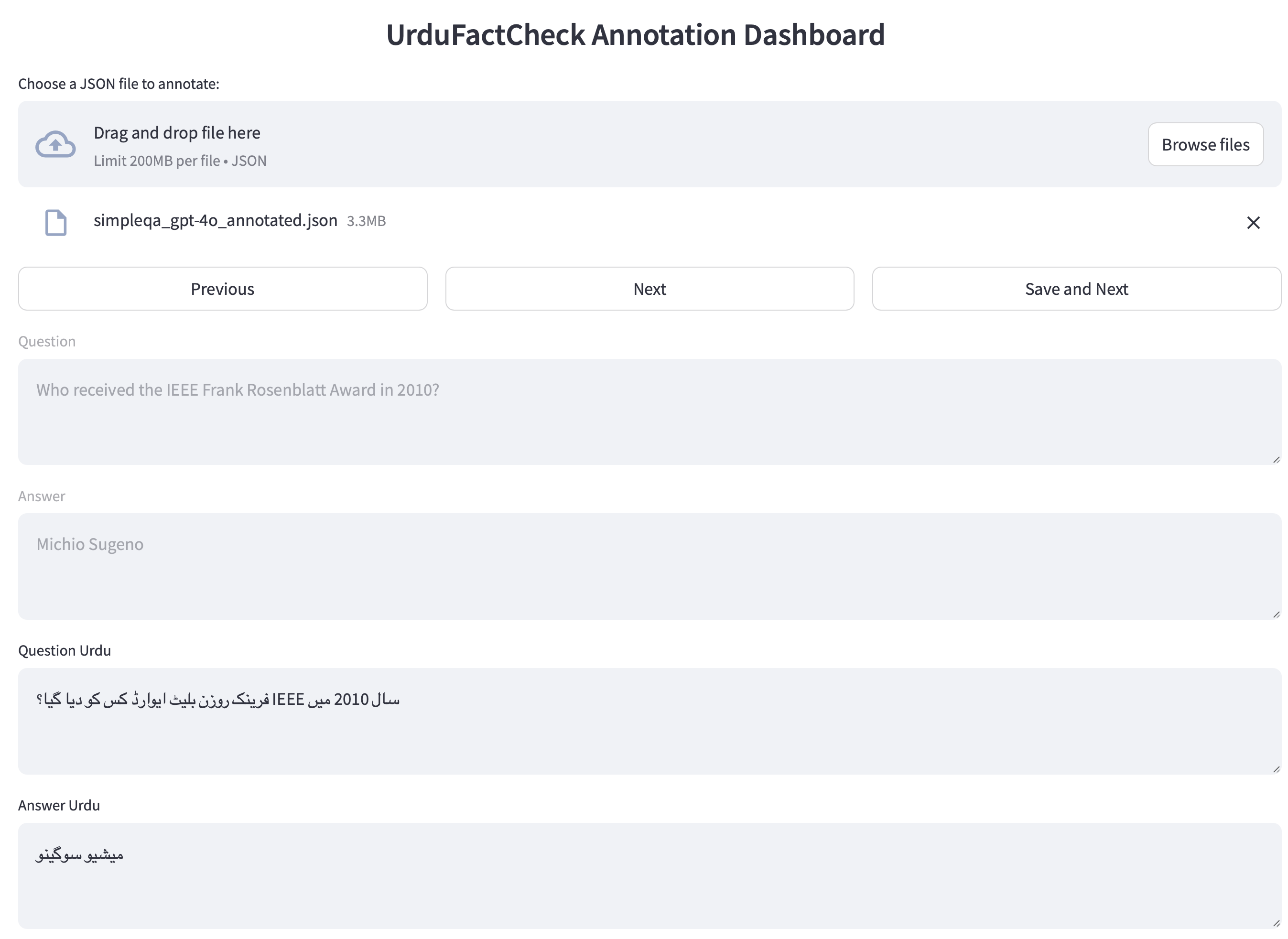}
  \caption{\urdufactcheck{} annotator dashboard built in Streamlit.}
  \label{fig:dashboard}
\end{figure*}

\section{\texorpdfstring{\urdufactcheck}{UrduFactCheck} Annotator Dashboard}
\label{app:annotator}

As shown in \autoref{fig:dashboard}, we developed a dedicated annotator dashboard to streamline dataset creation and quality assurance. 
Implemented in Streamlit, the dashboard offered expert annotators 
a simple interface to translate and review claim–label pairs, 
making it easier to ensure high-quality and consistent Urdu translations.

\section{\texorpdfstring{\urdufactcheck}{UrduFactCheck} Prompts}
\label{app:prompts}

This section presents the prompts that power the core modules of \urdufactcheck{}. 

\subsection*{\claimprocessor Prompt}

\begin{codebox}
\renewcommand{\baselinestretch}{1.2} 
{\fontsize{8.5pt}{10.5pt}\selectfont
\ttfamily
\raggedright
\hfill \foreignlanguage{urdu}{آپ کو ایک ایسا متن دیا گیا ہے جس میں علم کے دعوے شامل ہیں۔}\\

\hfill \foreignlanguage{urdu}{دعویٰ ایک بیان ہے جو کچھ سچ یا جھوٹ ہونے کا دعویٰ کرتا ہے،}\\

\hfill \foreignlanguage{urdu}{جس کی تصدیق انسانوں سے کی جا سکتی ہے۔ آپ کا کام یہ ہے}\\

\hfill \foreignlanguage{urdu}{کہ آپ دیے گئے متن میں سے ہر دعوے کو درست طریقے سے}\\

\hfill \foreignlanguage{urdu}{شناخت اور نکالیں۔ پھر، کسی بھی کورفرنس یعنی ضمیر یا}\\

\hfill \foreignlanguage{urdu}{دوسرے حوالہ دینے والے اظہار کو دعوے کی وضاحت کے لیے}\\

\hfill \foreignlanguage{urdu}{حل کریں۔ ہر دعویٰ مختصر یعنی} 15 \foreignlanguage{urdu}{الفاظ سے کم اور}\\

\hfill \foreignlanguage{urdu}{خود مختار ہونا چاہیے۔}\\

\hfill \foreignlanguage{urdu}{متن اردو میں دیا گیا ہے اور دعوے اردو میں نکالے جانے چاہئیں۔}\\

\hfill \foreignlanguage{urdu}{آپ کا جواب صرف نیچے دیے گئے فارمیٹ میں ہونا چاہیے۔}\\

\hfill \foreignlanguage{urdu}{اس کے علاوہ کوئی اور اضافی نوٹس یا وضاحت شامل نہ کریں۔}\\
~\\

~\hfill \foreignlanguage{urdu}{جواب کا فارمیٹ:}\\
\hfill\texttt{]}\\
\hfill\texttt{\}}\\
\hfill \foreignlanguage{urdu}{یقین دہانی کرائیں کہ دعویٰ} 15 \foreignlanguage{urdu}{الفاظ سے کم ہو} \texttt{: "claim"}\\

~ \hfill \foreignlanguage{urdu}{اور مکمل خیال فراہم کرے۔ کورفرنس کو دعوے کی}\\

~ \hfill \foreignlanguage{urdu}{وضاحت کے لیے حل کریں}\texttt{",}\\
\hfill\texttt{,\{}\\
\hfill\texttt{...}\\
\hfill\texttt{[}\\

~\hfill \foreignlanguage{urdu}{یہاں دو مثالیں دی گئی ہیں:}\\
~\hfill \{examples\}\\
~\\

~\hfill \foreignlanguage{urdu}{اب یہ مکمل کریں، صرف جواب کی شکل میں، کوئی اور الفاظ نہیں:}\\
~ \hfill \texttt{\{input\} :[text]}\\
~ \hfill \texttt{:[response]}
}
\end{codebox}
\captionof{lstlisting}{Our claim extraction prompt, which guides the identification of factual claims in Urdu text, with rules for co-reference resolution and concise, self-contained phrasing.}
\label{fig:claim-extraction-prompt}

\subsection{\querygenerator Prompt}

\begin{codebox}
\renewcommand{\baselinestretch}{1.2} 
{\fontsize{8.5pt}{10.5pt}\selectfont
\ttfamily
\raggedright
\hfill \foreignlanguage{urdu}{آپ ایک سوالات بنانے والے ہیں جو صارفین کو دیے گئے}\\

\hfill \foreignlanguage{urdu}{دعوے کو تلاش کے انجن کے ذریعے تصدیق کرنے میں مدد کرتے ہیں۔}\\

\hfill \foreignlanguage{urdu}{آپ کا بنیادی کام دو موثر اور شک انگیز تلاش کے انجن کے}\\

\hfill \foreignlanguage{urdu}{سوالات تیار کرنا ہے۔ یہ سوالات صارفین کو دیے گئے دعوے}\\

\hfill \foreignlanguage{urdu}{کی حقیقت کو تنقیدی طور پر جانچنے میں مدد فراہم کریں گے۔}\\

\hfill \foreignlanguage{urdu}{سوالات اردو میں ہونے چاہئیں اور سوالات اردو میں بنائے جائیں۔}\\

\hfill \foreignlanguage{urdu}{آپ کو صرف نیچے دیے گئے فارمیٹ میں جواب دینا ہوگا}\\

\hfill \foreignlanguage{urdu}{پائیتھون کی فہرست میں سوالات۔ براہ کرم اس فارمیٹ کی}\\

\hfill \foreignlanguage{urdu}{سختی سے پیروی کریں۔ کچھ اور واپس نہ کریں۔ اپنا جواب}\\

\hfill \texttt{'['} \foreignlanguage{urdu}{سے شروع کریں۔}\\

~\\

\hfill \texttt{['}\foreignlanguage{urdu}{سوال}\texttt{1', '}\foreignlanguage{urdu}{سوال}\texttt{2']} \foreignlanguage{urdu}{جواب کا فارمیٹ:} \\

~\hfill \foreignlanguage{urdu}{یہاں دو مثالیں دی گئی ہیں:}\\
~\hfill \{examples\}\\
~\\

~\hfill \foreignlanguage{urdu}{اب یہ مکمل کریں، صرف جواب کی شکل میں، کوئی اور الفاظ نہیں:}\\
~ \hfill \texttt{\{input\} :[claim]}\\
~ \hfill \texttt{:[response]}
}
\end{codebox}
\captionof{lstlisting}{Query generation prompt used by the \querygenerator{} module to create search engine queries for fact-checking claims. The prompt ensures queries are generated in Urdu and follow a specific format for effective claim verification.}
\label{fig:query-generation-prompt}

\subsection{\verifier Prompt}

\begin{codebox}
\renewcommand{\baselinestretch}{1.2} 
{\fontsize{8.5pt}{10.5pt}\selectfont
\ttfamily
\raggedright
\hfill \foreignlanguage{urdu}{آپ کو ایک ٹکڑا دیا گیا ہے۔ آپ کا کام یہ ہے کہ آپ یہ}\\

\hfill \foreignlanguage{urdu}{شناخت کریں کہ آیا دیے گئے متن میں کوئی حقیقت کی}\\

\hfill \foreignlanguage{urdu}{غلطیاں ہیں۔ جب آپ دیے گئے متن کی حقیقت کو پرکھ}\\

\hfill \foreignlanguage{urdu}{رہے ہوں، تو آپ ضرورت کے مطابق فراہم کردہ شواہد}\\

\hfill \foreignlanguage{urdu}{کا حوالہ دے سکتے ہیں۔ فراہم کردہ شواہد مددگار ہو سکتے ہیں۔}\\

\hfill \foreignlanguage{urdu}{بعض شواہد ایک دوسرے سے متضاد ہو سکتے ہیں۔ آپ کو}\\

\hfill \foreignlanguage{urdu}{شواہد کو احتیاط سے استعمال کرنا چاہیے جب آپ دیے گئے}\\

\hfill \foreignlanguage{urdu}{متن کی حقیقت کا اندازہ لگائیں۔}\\
~\\

\hfill \foreignlanguage{urdu}{جواب ایک ڈکشنری ہونی چاہیے جس میں چار کلیدیں ہوں}\\

\hfill \texttt{"reasoning" (}\foreignlanguage{urdu}{وجہ}\texttt{), "factuality" (}\foreignlanguage{urdu}{حقیقت}\texttt{)}\\

\hfill \foreignlanguage{urdu}{اور} \texttt{"error" (}\foreignlanguage{urdu}{غلطی}\texttt{)} \foreignlanguage{urdu}{اور} \texttt{"correction" (}\foreignlanguage{urdu}{تصحیح}\texttt{),}\\

\hfill \foreignlanguage{urdu}{جو بالترتیب آپ کی وجہ، یہ کہ آیا دیے گئے متن میں کوئی حقیقتی}\\

\hfill "Boolean True or False" \foreignlanguage{urdu}{غلطی ہے یا نہیں} \\

\hfill \foreignlanguage{urdu}{اور غلطی کی وضاحت، اور تصحیح فراہم کریں۔}\\

\hfill \foreignlanguage{urdu}{وجہ، غلطی اور تصحیح اردو میں ہونی چاہیے۔}\\
~\\

~\hfill \foreignlanguage{urdu}{جواب کا فارمیٹ:}\\
~\hfill \texttt{\}}\\
\foreignlanguage{urdu}{کیوں دی گئی عبارت حقیقت پر مبنی ہے یا نہیں؟} \texttt{:"reasoning"}\\

~ \hfill \foreignlanguage{urdu}{جب آپ یہ کہتے ہیں کہ کوئی چیز حقیقت پر مبنی نہیں ہے،}\\

~ \hfill \foreignlanguage{urdu}{تو آپ کو اپنے فیصلے کی حمایت کرنے کے لیے متعدد}\\

~ \hfill \foreignlanguage{urdu}{شواہد فراہم کرنے ہوں گے۔}\\
~ \hfill \foreignlanguage{urdu}{، ورنہ} \texttt{'None'} \foreignlanguage{urdu}{اگر عبارت حقیقت پر مبنی ہے تو} \texttt{:"error"}\\

~ \hfill \foreignlanguage{urdu}{غلطی کی وضاحت کریں۔}\\
~ \hfill \foreignlanguage{urdu}{اگر کوئی غلطی ہو تو تصحیح شدہ عبارت}\texttt{:"correction"}\\
~ \hfill \foreignlanguage{urdu}{ فراہم کریں۔}\\
~ \hfill \foreignlanguage{urdu}{اگر دی گئی عبارت حقیقت پر مبنی ،}\texttt{ True :"factuality"}\\
~ \hfill \texttt{False} \foreignlanguage{urdu}{ورنہ} \foreignlanguage{urdu}{ہے}\\
~ \hfill \texttt{\{} \\
~ \\

~\hfill \foreignlanguage{urdu}{اب یہ مکمل کریں، صرف جواب کی شکل میں، کوئی اور الفاظ نہیں:}\\
~ \hfill \texttt{\{claim\} :[claim]}\\
~ \hfill \texttt{\{evidence\} :[evidence]}\\
~ \hfill \texttt{:[response]}
}
\end{codebox}
\captionof{lstlisting}{Verification prompt used by the \verifier{} module to assess claim factuality based on retrieved evidence. The prompt ensures systematic evaluation and structured output with reasoning, error identification, and corrections in Urdu.}
\label{fig:verification-prompt}

\subsection*{Urdu to English Translator Prompt}

\begin{codebox}
\renewcommand{\baselinestretch}{1.2} 
{\fontsize{8.5pt}{10.5pt}\selectfont
\ttfamily
You are given a piece of text in Urdu. Your task is to translate it into English. The translation should be accurate and maintain the original meaning of the text. Please ensure that the translation is grammatically correct and coherent in English.\\
DO NOT RESPOND WITH ANYTHING ELSE. ADDING ANY OTHER EXTRA NOTES THAT VIOLATE THE RESPONSE FORMAT IS BANNED. \\

\{input\}
}
\end{codebox}
\captionof{lstlisting}{Prompt for translating Urdu text into English.}
\label{fig:ue-prompt}

\subsection{English to Urdu Translator Prompt}

\begin{codebox}
\renewcommand{\baselinestretch}{1.2} 
{\fontsize{8.5pt}{10.5pt}\selectfont
\ttfamily
You are given a piece of text in English. Your task is to translate it into Urdu. The translation should be accurate and maintain the original meaning of the text. Please ensure that the translation is grammatically correct and coherent in Urdu.
DO NOT RESPOND WITH ANYTHING ELSE. ADDING ANY OTHER EXTRA NOTES THAT VIOLATE THE RESPONSE FORMAT IS BANNED.\\

\{input\}
}
\end{codebox}
\captionof{lstlisting}{Prompt for translating English text into Urdu.}
\label{fig:eu-prompt}

\end{document}